\documentclass[12pt]{extarticle}

\usepackage[margin=1in]{geometry}
\usepackage[T1]{fontenc}
\IfFileExists{lmodern.sty}{\usepackage{lmodern}}{}
\usepackage{amsmath,amssymb,mathtools}
\usepackage{booktabs,array,longtable}
\usepackage{graphicx}
\usepackage{xcolor}
\usepackage{tikz}
\usetikzlibrary{arrows.meta,positioning,fit,backgrounds,calc,shapes.geometric}
\usepackage{enumitem}
\usepackage[round]{natbib}
\usepackage{hyperref}
\usepackage{url}
\usepackage{listings}
\IfFileExists{microtype.sty}{%
  \usepackage{microtype}%
  \IfFileExists{lmodern.sty}{}{\microtypesetup{expansion=false}}%
}{}
\usepackage{caption}
\usepackage{float}

\hypersetup{
  colorlinks=true,
  linkcolor=blue!60!black,
  citecolor=blue!60!black,
  urlcolor=blue!60!black
}

\definecolor{detgreen}{RGB}{46,125,50}
\definecolor{stochamber}{RGB}{198,108,0}
\definecolor{sysred}{RGB}{183,28,28}
\definecolor{agentblue}{RGB}{25,86,155}
\definecolor{toolpurple}{RGB}{106,27,154}
\definecolor{softgray}{RGB}{248,248,248}

\newcolumntype{P}[1]{>{\raggedright\arraybackslash}p{#1}}

\lstdefinestyle{pythonstyle}{
  language=Python,
  basicstyle=\ttfamily\small,
  numbers=left,
  numberstyle=\tiny,
  stepnumber=1,
  numbersep=6pt,
  showstringspaces=false,
  breaklines=true,
  columns=fullflexible,
  frame=single,
  tabsize=4
}

\lstdefinestyle{shellstyle}{
  language=bash,
  basicstyle=\ttfamily\small,
  numbers=none,
  showstringspaces=false,
  breaklines=true,
  columns=fullflexible,
  frame=single,
  tabsize=2
}

\newcommand{\Vocab}{\mathcal{V}}
\newcommand{\Vocabk}{\mathcal{V}_k}
\newcommand{\Vocabp}{\mathcal{V}_p}

\newcommand{\R}{\mathbb{R}}

\newcommand{\softmax}{\operatorname{softmax}}
\newcommand{\argmax}{\operatorname*{arg\,max}}
\newcommand{\append}{\mathbin{\Vert}}
\newcommand{\Categorical}{\operatorname{Categorical}}
\newcommand{\Parse}{\operatorname{Parse}}
\newcommand{\Update}{\operatorname{Update}}

\renewcommand{\arraystretch}{1.15}
\title{The Token Not Taken: Sampling, State, and the Stochasticity of AI Agents\thanks{\copyright~2026 Muhammad Zia Hydari and Raja Iqbal.}}
\author{%
  Muhammad Zia Hydari\\
  University of Pittsburgh\\
  \texttt{hydari@alum.mit.edu}\\
  ORCID: 0000-0003-4522-326X
  \and
  Raja Iqbal\\
  Ejento AI Inc.\\
  \texttt{raja@ejento.ai}
}

\begin{document}
\maketitle

\begin{abstract}
Agentic AI systems can behave differently across runs: the same request may produce a different plan, a different tool call, a different code edit, or a different final answer. Such variability arises from several layers that are often conflated. At the core of many current agents is a foundation model, a large pretrained model adaptable to many downstream tasks, embedded in an orchestration loop that plans, calls tools, observes results, and updates state. One explicit intrinsic source of variability in such systems is token generation: the model computes scores over possible next tokens, the scores are converted into probabilities, and a decoder may sample tokens using a pseudo-random number generator. A small sampled token difference can then cascade downstream into a different tool call, code path, search query, or agent state. Other sources of variability are extrinsic to token sampling, including changing environments, live data, serving infrastructure, batch effects, and numerical details. By separating these layers, this tutorial clarifies what it means to call agentic AI systems stochastic, when such variability can be reproduced under matched conditions, and why deterministic execution need not imply identical behavior in deployed settings.
\end{abstract}

\vspace{0.5em}
\noindent\textbf{Keywords:} Agentic AI, stochasticity, foundation models, decoding, reproducibility.

\section{Introduction}
\label{sec:agentic-phenomenon}

Generative AI systems are widely understood to be stochastic: they can produce different outputs across runs, even when given the same visible input. A chatbot may answer the same question with different wording, different examples, or a different ordering of ideas. In many ordinary uses, this variability is expected and even useful.

An \emph{agentic AI system} is a generative AI application that uses a foundation model not only to produce text, but also to plan, call tools, observe results, update state, and take further actions. This makes variability more consequential. A difference that appears harmless in a chatbot response may become a different search query, tool call, code edit, recommendation, escalation, or stopping decision in an agentic system. Managers, product owners, instructors, and researchers are therefore often concerned that an AI agent may do something different across runs, even when the visible request appears to be the same.

\begin{figure}[H]
\centering
\resizebox{0.92\textwidth}{!}{%
\begin{tikzpicture}[node distance=12mm and 16mm]
  \tikzset{
    comp/.style  = {rounded corners=3pt, align=center, font=\small,
                    minimum height=11mm, line width=0.6pt},
    fm/.style    = {comp, draw=detgreen, fill=detgreen!8,  text width=27mm},
    dec/.style   = {comp, draw=stochamber, fill=stochamber!12, text width=20mm},
    sat/.style   = {comp, draw=agentblue, fill=agentblue!7, text width=33mm},
    toolc/.style = {comp, draw=toolpurple, fill=toolpurple!8, text width=34mm},
    envc/.style  = {comp, draw=sysred, fill=sysred!7, text width=32mm},
    link/.style  = {-{Stealth[length=2.2mm]}, black!70, line width=0.5pt},
    bilink/.style= {{Stealth[length=2.2mm]}-{Stealth[length=2.2mm]}, black!70, line width=0.5pt}
  }

  \node[fm] (fm) {Foundation\\model (LLM)};
  \node[dec, right=9mm of fm] (dec) {Decoder};

  \begin{scope}[on background layer]
    \node[draw=black!55, dashed, rounded corners=6pt, fill=black!2,
          fit=(fm)(dec), inner sep=7mm] (runtime) {};
  \end{scope}
  \node[font=\footnotesize\itshape, black!60, anchor=north]
        at ([yshift=-1.2mm]runtime.north) {agent runtime / orchestration loop};

  \node[sat, above=13mm of runtime] (instr) {System\\instructions,\\role, and policies};
  \node[sat, left=14mm of runtime]  (mem)   {Memory \& retrieval\\(history, RAG)};
  \node[sat, below=13mm of runtime] (guard) {Guardrails \&\\output validation};
  \node[toolc, right=18mm of runtime] (tools) {Tools:\\search, code, database, APIs};
  \node[envc, below=13mm of tools]  (env)   {External\\environment: web,\\files, clock, users};

  \draw[link]   (instr.south) -- (instr.south |- runtime.north);
  \draw[bilink] (mem.east)    -- (mem.east -| runtime.west);
  \draw[link]   (guard.north |- runtime.south) -- (guard.north);
  \draw[bilink] (runtime.east) -- (tools.west);
  \draw[bilink] (tools.south)  -- (env.north);

  \node[stochamber, font=\scriptsize, below=1mm of dec]
        {intrinsic sampling enters here};
\end{tikzpicture}}
\caption{Anatomy of an agentic AI system. A foundation model and decoder sit inside an orchestration loop, surrounded by instructions, memory and retrieval (retrieval-augmented generation, RAG), tools, guardrails, and an external environment.}
\label{fig:agent-architecture}
\end{figure}
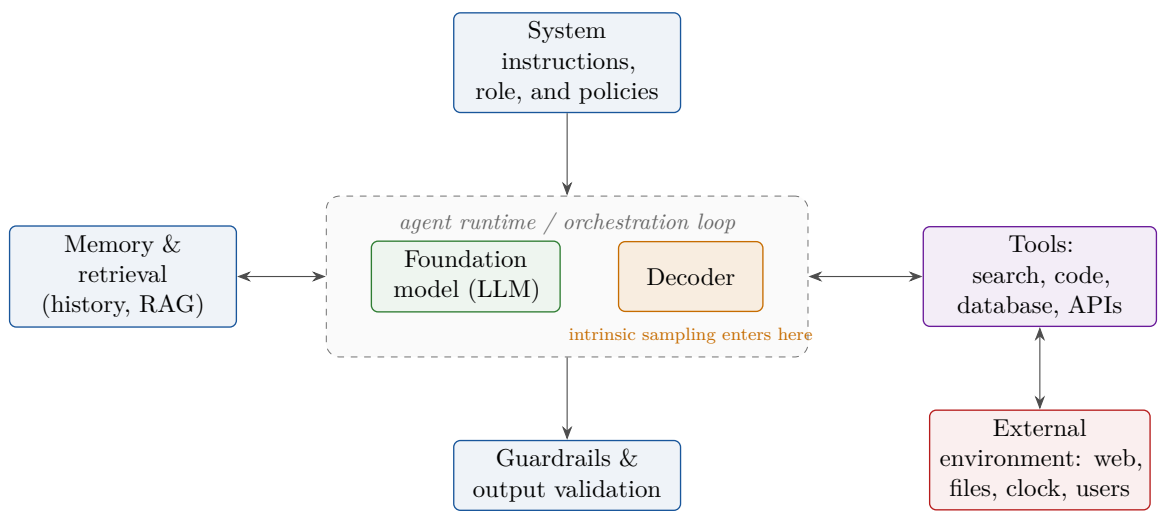

At the core of many current agentic AI systems is a foundation model~\citep{bommasani2021foundation}, embedded in an orchestration loop with instructions, tools, memory or state, guardrails, and mechanisms for acting on and observing an external environment \citep{openai-agents, openai-agents-guide, anthropic-tool-use, langchain-agents}.\footnote{LangChain is an open-source framework for building applications around language models; it supplies the orchestration loop, tool interfaces, and memory components sketched here. The OpenAI Agents software development kit (SDK) and Anthropic's tool-use interface play comparable roles.} Figure~\ref{fig:agent-architecture} sketches how these pieces fit together.

This tutorial separates variability across levels of analysis. At the computational level, a model call produces raw scores over possible next tokens, converts those scores into probabilities, and may sample from those probabilities. At the model-output level, this may appear as a different sentence, JavaScript Object Notation (JSON) object, code fragment, or plan. At the agent-action level, it may appear as a different tool call, search query, file inspection, code edit, or stopping decision. At the workflow level, it may appear as a different customer-routing decision, refund recommendation, report, compliance exposure, or pull-request outcome. Keeping these levels separate is important: token-level variation is not itself a business outcome, but it can be amplified through the agent loop into actions and outcomes that users observe.

The following examples are intentionally simple. Their purpose is not to show that agentic systems exist, but to establish the level at which variability matters for this tutorial. The relevant phenomenon is not merely that two generated texts differ in wording. Rather, two runs of the same agentic system may diverge in the actions they take and the outcomes they produce.

\begin{enumerate}[leftmargin=*,label=\textbf{Example \arabic*.}]
\item \textbf{Coding agents.} GitHub documents a Copilot cloud agent that can be assigned tasks, work in a repository, and create code changes in a pull request workflow \citep{github-copilot-cloud}. OpenAI describes Codex as a coding agent that can help with software development tasks \citep{openai-codex}. For the same issue, a coding agent might inspect different files, generate different tests, or implement a fix with a different design.

\item \textbf{Customer support agents.} A support agent may have tools for customer lookup, refund eligibility, issuing credits, and escalation to a human. For the same customer message, one run may ask a clarifying question, another may check account status first, and another may recommend escalation. The system's variability is observed as a different action in a business process. This example will serve as the running example in later sections.

\item \textbf{Research and analysis agents.} A data-analysis agent might generate Python code, execute it, inspect an error message, revise the code, and summarize the result. A research agent might issue search queries, retrieve documents, read results, and synthesize an answer. Tool-using systems explicitly support actions such as fetching data, running code, and calling application programming interfaces (APIs) \citep{anthropic-tool-use,openai-tools}.
\end{enumerate}

These examples put the phenomenon at the level at which users experience it: variation in plans, actions, observations, and outcomes. Much of the intrinsic variation in such systems begins in foundation-model token generation, but the agent loop can amplify a small model-output difference into a different trajectory.

This tutorial is written for information systems (IS) researchers, doctoral students, instructors, MBA students, and technically oriented managers who need a precise account of why agentic AI systems can behave differently across runs. The tutorial assumes familiarity with AI applications and organizational use cases, but it does not assume specialized training in machine-learning infrastructure. Its aim is to connect several levels of analysis: the organizational phenomenon of variable agent behavior, the token-level mechanics of foundation-model generation, and the systems-level sources of variation introduced by tools, environments, and serving infrastructure.

The tutorial also provides technical background for a related managerial discussion of agentic technical debt and stochastic tax, which treats stochastic agent behavior as an operating and governance burden in enterprise deployments \citep{hydariInPressGoverning}.

The rest of the tutorial proceeds as follows. Section~\ref{sec:typology} presents a typology of variability in agentic AI systems. Section~\ref{sec:agent-model} gives a compact state-machine model of an agent. Sections~\ref{sec:model-call} and~\ref{sec:decoding} explain what happens inside a single foundation-model call: vocabulary, hidden representations, logits, probabilities, and decoding. Section~\ref{sec:amplification} explains how token-level variation becomes agent-level variation. Section~\ref{sec:not-mean} clarifies common misconceptions, including reproducibility, reasoning paths, and graphics processing unit (GPU) nondeterminism. Section~\ref{sec:tracing-governance} draws out implications for tracing and governing agentic variability. Appendix~\ref{app:notation} collects the notation, and the remaining appendices provide runnable code demonstrations.

\paragraph{Reading path.}
Readers primarily interested in the conceptual argument may skim the detailed equations and numerical examples in Sections~\ref{sec:model-call} and~\ref{sec:decoding} on a first pass. The prose provides the main thread: model contexts are converted into raw next-token scores, scores are converted into probabilities, and stochastic decoding samples from those probabilities. The equations and toy examples are included for precision, reproducibility, and classroom use.

\section{A typology of variability in agentic AI systems}
\label{sec:typology}

A useful way to organize agentic variability is to distinguish the layer at which variation enters. Table~\ref{tab:typology} presents a working typology. The ordering is conceptual rather than chronological: it begins with the mechanism most directly associated with stochastic model output, namely sampled decoding, and then moves outward to model serving, context construction, orchestration, tool and environment responses, and business-process consequences.

Throughout the tutorial, we distinguish sources of variation from manifestations and consequences. A sampled token is a source of variation at the decoding layer. A different generated JSON object, plan sentence, or code fragment is a manifestation at the model-output layer. A different tool call, database query, file inspection, or code edit is a manifestation at the agent-action layer. A different refund decision, customer routing, compliance exposure, or pull request outcome is a business-process consequence. The levels are connected, but they should not be conflated.

{\footnotesize
\setlength{\tabcolsep}{3pt}%
\renewcommand{\arraystretch}{1.08}%
\begin{longtable}{P{0.14\textwidth}P{0.18\textwidth}P{0.16\textwidth}P{0.23\textwidth}P{0.19\textwidth}}
\caption{A working typology of variability in agentic AI systems. The table distinguishes where variation enters, what varies directly, how the variation appears in an agent trace, and what kind of control or replay is possible.}\label{tab:typology}\\
\toprule
\textbf{Layer} & \textbf{Source or mechanism} & \textbf{What varies directly} & \textbf{Manifestation in an agent trace} & \textbf{Control or replay strategy} \\
\midrule
\endfirsthead
\toprule
\textbf{Layer} & \textbf{Source or mechanism} & \textbf{What varies directly} & \textbf{Manifestation in an agent trace} & \textbf{Control or replay strategy} \\
\midrule
\endhead
Decoding / token sampling & Temperature, top-\(k\), top-\(p\), pseudo-random draws & Next token or generated output sequence & Different wording, JSON field, tool-call string, code token, or plan step & Often reproducible with fixed model, fixed context, fixed decoder, and synchronized pseudo-RNG state. \\
Model computation and serving & Batch composition, low-level numerical routines, precision, routing, model version, or mixture-of-experts (MoE) load effects & Logits or next-token probability vector & Same apparent prompt may yield slightly different token probabilities, even before sampling & A seed usually cannot remove this in managed services; self-hosted systems may control some numerical routines, shapes, versions, and batching. \\
State and context construction & Retrieval, memory, prompt assembly, tool availability, or policy context & Serialized model context received by the model & The model call is not actually repeated under the same conditions & Freeze or replay the constructed context, retrieval outputs, memory state, tool list, and policies. \\
Parsing and orchestration & Parser rules, schema validation, retries, guardrails, fallback paths, or timeouts & Parsed action or control path & Different tool call, retry, code edit, escalation, or stop/continue decision & Sometimes controllable if application state, parser version, validation rules, and timing are controlled. \\
Tool and environment response & Live databases, APIs, webpages, files, clock time, or user responses & Observation returned to the agent & Same action receives a different tool result or external observation & Use snapshots, mocks, recorded tool outputs, or replay environments. \\
Workflow or business consequence & Accumulation of upstream variation through an organizational process & Business-relevant outcome & Different customer routing, refund decision, report, code patch, compliance exposure, or user experience & Not controlled directly by a seed; must be traced, governed, and measured at the trajectory or process level. \\
\bottomrule
\end{longtable}
}

The typology makes two points central to the rest of the tutorial. First, token sampling is the clearest intrinsic source of stochasticity in generation, because a decoder may draw tokens from a probability distribution using a pseudo-random number generator (PRNG). Second, deployed agents also vary for reasons outside token sampling. A seed cannot make a webpage remain unchanged, keep a database row fixed, prevent an API timeout, or force a managed inference service to use the same hidden serving conditions. Reproducibility claims therefore require attention to the whole agent trajectory, not only to the model call.

\section{A compact model of agentic stochasticity}
\label{sec:agent-model}

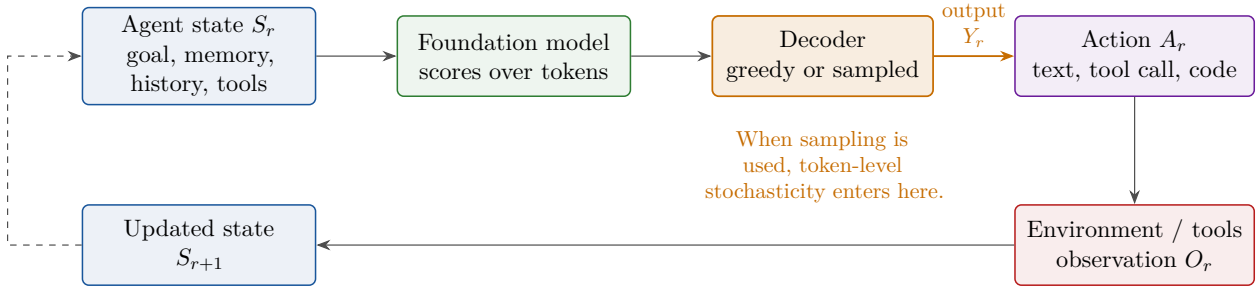
\begin{figure}[H]
\centering
\resizebox{\textwidth}{!}{%
\begin{tikzpicture}[node distance=13mm]
  \tikzset{
    statebox/.style={draw=agentblue, rounded corners=3pt, line width=0.7pt, fill=agentblue!8, align=center, text width=35mm, minimum height=13mm, font=\small},
    modelbox/.style={draw=detgreen, rounded corners=3pt, line width=0.7pt, fill=detgreen!8, align=center, text width=35mm, minimum height=13mm, font=\small},
    samplebox/.style={draw=stochamber, rounded corners=3pt, line width=0.8pt, fill=stochamber!12, align=center, text width=33mm, minimum height=13mm, font=\small},
    actionbox/.style={draw=toolpurple, rounded corners=3pt, line width=0.7pt, fill=toolpurple!8, align=center, text width=36mm, minimum height=13mm, font=\small},
    envbox/.style={draw=sysred, rounded corners=3pt, line width=0.7pt, fill=sysred!8, align=center, text width=36mm, minimum height=13mm, font=\small},
    flow/.style={-{Stealth[length=2.3mm]}, black!70, line width=0.55pt},
    stochastic/.style={-{Stealth[length=2.3mm]}, stochamber, line width=0.85pt},
    feedback/.style={-{Stealth[length=2.3mm]}, black!70, dashed, line width=0.55pt}
  }

  \node[statebox] (state) {Agent state $S_r$\\goal, memory, history, tools};
  \node[modelbox, right=of state] (model) {Foundation model\\scores over tokens};
  \node[samplebox, right=of model] (decode) {Decoder\\greedy or sampled};
  \node[actionbox, right=of decode] (action) {Action $A_r$\\text, tool call, code};
  \node[statebox, below=16mm of state] (nextstate) {Updated state\\$S_{r+1}$};
  \node[envbox] (env) at (nextstate -| action) {Environment / tools\\observation $O_r$};

  \draw[flow] (state) -- (model);
  \draw[flow] (model) -- (decode);
  \draw[stochastic] (decode) -- node[above, align=center, font=\footnotesize, stochamber] {output\\$Y_r$} (action);
  \draw[flow] (action) -- (env);
  \draw[flow] (env.west) -- ++(-25mm,0) |- (nextstate.east);
  \draw[feedback] (nextstate.west) -- ++(-12mm,0) |- (state.west);

  \node[stochamber, align=center, font=\footnotesize, below=4mm of decode, text width=39mm]
    {When sampling is used, token-level stochasticity enters here.};
\end{tikzpicture}}
\caption{Agentic stochasticity. The agent repeatedly turns model outputs into actions and observations. A small sampled token difference can change a tool call, code edit, or subsequent state.}
\label{fig:agent-loop}
\end{figure}

To precisely identify where stochasticity enters, we can formalize the agent's operation as a state machine. Let an agentic system maintain a state
\[
S_r = (G, C_r, M_r, \mathsf{Tools}_r, E_r),
\]
where \(G\) is the user's goal, \(C_r\) is the conversation history available at agent round \(r\), \(M_r\) is memory or retrieved material, \(\mathsf{Tools}_r\) is the currently available tool set, and \(E_r\) contains environment information such as time, files, database contents, web observations, API responses, or user interactions.\footnote{A \emph{state} is the information the system has available at a given moment. For an agent, the state may include visible conversation history, hidden system instructions, retrieved material, memory, tool definitions, tool outputs, and external observations. This is broader than the model's immediate \emph{context}. In this tutorial, the model context refers to the serialized input sequence actually passed to the foundation model for a particular call. It may include some or all of the conversation history, retrieved material, tool definitions, prior tool outputs, and instructions, depending on how the agent constructs the prompt.}

In the running customer-support example, \(G\) might be the user's goal of resolving a refund dispute, \(C_r\) the conversation so far, \(M_r\) retrieved refund policy and account history, \(\mathsf{Tools}_r\) the available lookup, refund, credit, and escalation tools, and \(E_r\) the current state of the order database, clock, and support queue.

At round \(r\), the agent constructs a model context from this broader state. We write this serialized input as \(c_{r,1}\), the initial context for the token-generation loop in that model call. The foundation model then generates tokens sequentially, indexing \(n\) over token-generation steps. The model and decoder produce an output sequence
\[
Y_r = (o_{r,1}, o_{r,2}, \ldots, o_{r,L_r}),
\]
where each \(o_{r,n}\) is a generated token. The output sequence may be ordinary prose, code, JSON, a tool call, a plan, or a structured response. The agent then parses this output into an action:
\[
A_r = \Parse(Y_r).
\]

The environment/tool response map \(\mathcal{E}\) represents the external system that executes the agent's action and returns what the agent observes. The observation \(O_r\) depends not only on the action \(A_r\), but also on the current environment state \(E_r\). This state may include deterministic tools, live databases, web services, files, clocks, application programming interfaces (APIs), or human-facing systems whose contents may change over time. Thus,
\[
O_r = \mathcal{E}(A_r; E_r).
\]

The agent updates its state:
\[
S_{r+1}=\Update(S_r,Y_r,A_r,O_r).
\]
This gives the agentic loop:
\[
S_r \rightarrow Y_r \rightarrow A_r \rightarrow O_r \rightarrow S_{r+1}.
\]

The key intrinsic stochastic step is the generation of \(Y_r\). Before the pseudo-random stream is initialized or seeded, the output is naturally described as a sample from a distribution,
\[
Y_r \sim D_\lambda\big(P_\theta(\cdot \mid S_r)\big),
\]
where \(P_\theta(\cdot \mid S_r)\) denotes the foundation model's token probabilities conditioned on the current agent state, with the serialization of \(S_r\) into the initial context \(c_{r,1}\) left implicit, and \(D_\lambda\) denotes the decoding procedure with parameters \(\lambda\). Once the pseudo-random number generator state \(R_r\) is also fixed, together with the model weights, the model context, the decoding procedure, and the numerical execution environment, the same step is a deterministic function of its inputs,
\[
Y_r = D_\lambda\big(P_\theta(\cdot \mid S_r), R_r\big).
\]
This pair of expressions states the thesis precisely: the decoder samples, yet a run with a fixed pseudo-random state is reproducible.\footnote{This is the same distinction used in simulation. A Monte Carlo simulation can be described probabilistically, but a particular run with a fixed seed is fully reproducible if the program and numerical environment are controlled.}

Figure~\ref{fig:agent-loop} summarizes the loop. The token-level stochasticity enters inside the model call, but it becomes consequential when the output is parsed as an action. The typology in Table~\ref{tab:typology} reminds us that this intrinsic sampling step is only one source of variation in the full agentic system.

\section{What a foundation model computes at a single token step}
\label{sec:model-call}

To understand where intrinsic stochasticity enters, we zoom in on one foundation-model call. For the large language model (LLM)-based agents that dominate current practice, text generation is usually \emph{autoregressive}: the model generates one token at a time, and each generated token becomes part of the model context for the next token.\footnote{A token is a unit used by the model's tokenizer. It may be a word, part of a word, punctuation, whitespace, or a byte-level fragment. The model does not usually operate directly on English words. This tutorial focuses on the autoregressive LLM-based agents that dominate current practice. Other generative architectures may require different mechanics, but the distinction between model-level variation, system-level variation, and agent-level consequences remains useful.}

\subsection{The next-token choice is over a fixed vocabulary}

A tokenizer maps text into token IDs. Let the model's fixed vocabulary be
\[
\Vocab = \{\tau_1,\tau_2,\ldots,\tau_K\}.
\]
Here \(K=|\Vocab|\) is the vocabulary size. The model is not choosing the immediate next token from all possible sentences, paragraphs, or future continuations. It is assigning a score to each token in a finite vocabulary, typically on the order of tens of thousands or hundreds of thousands of tokens, depending on the tokenizer.

The combinatorial explosion appears over multi-token continuations. If a model has \(K\) possible next tokens at each step and generates \(L\) future tokens, then there are up to
\[
K^L
\]
possible token sequences of length \(L\). With \(K=50{,}000\) and \(L=20\), this is
\[
50{,}000^{20} \approx 9.5\times 10^{93}.
\]
The immediate next-token choice is finite, but the space of possible continuations grows exponentially in length. While the exact numbers vary, the order-of-magnitude lesson remains: even short continuations create an astronomically large space of possible token paths.

\subsection{From context tokens to raw next-token scores}

\begin{figure}[H]
\centering
\resizebox{\textwidth}{!}{%
\begin{tikzpicture}[
  node distance=10mm,
  flow/.style={-{Stealth[length=2.2mm]}, line width=0.6pt, black!70},
  stage/.style={rounded corners=3pt, line width=0.7pt, align=center,
                inner xsep=7pt, inner ysep=6pt, minimum height=16mm,
                font=\small}
]

\node[stage, draw=agentblue, fill=agentblue!7] (context)
  {$\displaystyle \underbrace{x_1,\ldots,x_m}_{\substack{\text{tokens already}\\[-1pt]\text{in the model context}}}$};

\node[stage, draw=detgreen, fill=detgreen!8, right=of context] (scores)
  {$\displaystyle \underbrace{\begin{array}{c}
      \tau_1,\ldots,\tau_K\\[-1pt]
      z_1,\ldots,z_K
    \end{array}}_{\substack{\text{candidate next tokens}\\[-1pt]\text{and raw scores}}}$};

\node[stage, draw=stochamber, fill=stochamber!10, right=of scores] (probs)
  {$\displaystyle \underbrace{p_T(\tau_1\mid c_n),\ldots,p_T(\tau_K\mid c_n)}_{\substack{\text{probabilities over}\\[-1pt]\text{candidate next tokens}}}$};

\node[stage, draw=toolpurple, fill=toolpurple!8, right=of probs] (selected)
  {$\displaystyle \underbrace{o_n}_{\substack{\text{selected}\\[-1pt]\text{next token}}}$};

\draw[flow] (context) -- (scores);
\draw[flow] (scores) -- (probs);
\draw[flow] (probs) -- (selected);

\end{tikzpicture}}
\caption{From context tokens to a selected next token. The tokens already in the model context are used to assign raw scores to candidate next tokens in the vocabulary. These scores are converted into probabilities, and the decoder selects the next token.}
\label{fig:context-to-next-token}
\end{figure}
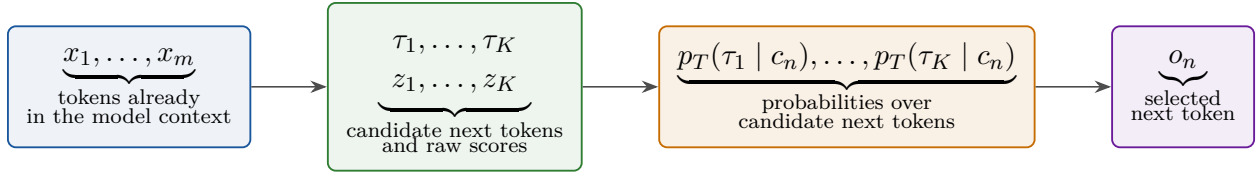

Suppose the current model context contains \(m\) tokens,
\[
x_1,x_2,\ldots,x_m.
\]
These are not the candidate next tokens. They are the tokens already available to the model before the next token is chosen. They may include the user's prompt, system instructions, previous conversation, retrieved material, tool outputs, and, during generation, tokens already generated in the current response. Given this existing context, the next-token problem is to assign a raw score to each candidate token in the vocabulary
\[
\Vocab=\{\tau_1,\tau_2,\ldots,\tau_K\}.
\]
The decoder then uses the resulting scores or probabilities to select the next output token. Figure~\ref{fig:context-to-next-token} summarizes the flow from existing context tokens to the selected next token. The probability notation in the figure is defined formally in the softmax subsection below.

The transformer maps the sequence of \(m\) model-context tokens into a matrix of hidden representations,
\[
H=f_\theta(x_1,x_2,\ldots,x_m)\in\R^{m\times d}.
\]
This equation says that the model, with parameters \(\theta\), applies its transformer layers to the input sequence and returns a matrix \(H\). The matrix has \(m\) rows because there are \(m\) token positions in the current model context. It has \(d\) columns because each token position is represented internally by a vector of \(d\) numerical components. The number \(d\) is called the hidden dimension. It is the width of the model's internal representation at each position, not the number of vocabulary tokens.

The notation \(H_{i,:}\) means row \(i\) of the matrix \(H\), with all columns included. Thus, \(H_{1,:}\) is the hidden representation at the first token position, \(H_{2,:}\) is the hidden representation at the second token position, and \(H_{m,:}\) is the hidden representation at the final position in the current model context. The colon means ``all columns''.

For next-token prediction in an autoregressive language model, the final-position representation is the one used to compute the next-token scores. We write it as
\[
h_m=(H_{m,:})^\top\in\R^d.
\]
Here \(H_{m,:}\) is a row vector of length \(d\). The transpose symbol \((\cdot)^\top\) turns it into a column vector, which is convenient for the linear map below. The vector \(h_m\) is therefore the final-position hidden representation used to predict the next token. Through causal attention, it can incorporate information from earlier positions in the model context.

The model now needs to turn the final-position hidden vector \(h_m\) into one raw score for every possible next token in the vocabulary. To do this, it uses an output projection matrix,
\[
W_U\in\R^{K\times d}.
\]
This matrix has \(K\) rows, one for each vocabulary token, and \(d\) columns, matching the dimension of \(h_m\). The \(i\)th row of \(W_U\), written \(w_i^\top\), is the learned output-side vector associated with candidate next token \(\tau_i\). These row vectors are part of the model's learned parameters: during training, they are adjusted so that tokens that should be likely in a given model context receive higher scores. At inference time, these rows are no longer updated by learning. For a fixed model checkpoint, each vocabulary token has a fixed output-side vector, while the hidden vector \(h_m\) changes with the current model context.\footnote{Some architectures tie the output projection matrix to the input embedding matrix, so the same learned token vectors are used on both the input and output sides. Other architectures use a separate output projection. This architectural detail does not change the role of \(W_U\) here: it maps the final hidden vector to one raw score per vocabulary token.} Thus, the same candidate-token vector can receive a high score in one context and a low score in another because the context-dependent vector \(h_m\) changes.

The raw score for candidate token \(\tau_i\) is computed by comparing the candidate token's output-side vector \(w_i\) with the final hidden vector \(h_m\):
\[
z_i=w_i^\top h_m+b_i.
\]
The dot product \(w_i^\top h_m\) is large when the final hidden vector is well aligned with the learned vector for token \(\tau_i\), and smaller when it is not. The term \(b_i\) is a learned bias for that token.

Stacking this calculation for all \(K\) candidate tokens gives the vector equation
\[
z=W_Uh_m+b.
\]
The full hidden-state matrix \(H\) contains one hidden representation for each context position. For next-token generation, however, we need the scores for the token that follows the entire current model context. Therefore, we use the final-position hidden vector \(h_m=(H_{m,:})^\top\). If one wanted scores for every position in the sequence, one could apply the output projection to every row of \(H\), producing an \(m\times K\) matrix of scores. For next-token generation, the relevant row is the final one.

Since \(W_U\) has \(K\) rows and \(d\) columns, and \(h_m\) has \(d\) entries, the product \(W_Uh_m\) has \(K\) entries. Adding \(b\in\R^K\) produces
\[
z=[z_1,z_2,\ldots,z_K]^\top\in\R^K,
\]
with one raw score for each possible next token in the vocabulary.

The vector \(z\) contains the model's raw scores for all candidate next tokens. These scores are not yet probabilities. They can be positive, negative, or zero, and they need not sum to one. In neural-network usage, this raw score vector is commonly called the vector of logits. The next step is to convert these raw scores into probabilities using the softmax function.

\paragraph{A toy numerical example.}
To make the notation concrete, suppose the current model context contains \(m=3\) tokens:
\[
x_1=\texttt{refund},\qquad
x_2=\texttt{order},\qquad
x_3=\texttt{please}.
\]
These are the tokens already in the context. They are not the candidate next tokens. Suppose also that the model's hidden dimension is \(d=2\), so each token position is represented by two internal numerical features. After the transformer processes the three context tokens, suppose it returns the hidden-representation matrix
\[
H
=
\begin{bmatrix}
0.2 & 0.1\\
0.4 & 0.7\\
1.0 & 0.5
\end{bmatrix}
\in\R^{3\times 2}.
\]
The three rows correspond to the three context positions, and the two columns correspond to the two hidden features. The final row is
\[
H_{3,:}=[1.0,\;0.5].
\]
Thus the final-position hidden vector is
\[
h_3=(H_{3,:})^\top
=
\begin{bmatrix}
1.0\\
0.5
\end{bmatrix}
\in\R^2.
\]
This is the vector the model uses to score candidate next tokens.

Now suppose the toy vocabulary has \(K=4\) candidate next tokens:
\[
\Vocab
=
\{\tau_1,\tau_2,\tau_3,\tau_4\}.
\]
Let
\[
\begin{aligned}
\tau_1&=\texttt{lookup\_order}, &
\tau_2&=\texttt{escalate\_case},\\
\tau_3&=\texttt{ask\_clarify}, &
\tau_4&=\texttt{deny\_request}.
\end{aligned}
\]
The output projection matrix and bias vector are
\[
W_U
=
\begin{bmatrix}
2.0 & 1.0\\
1.0 & 0.5\\
0.5 & 1.0\\
-1.0 & 0.0
\end{bmatrix},
\qquad
b
=
\begin{bmatrix}
0.0\\
0.1\\
0.0\\
0.0
\end{bmatrix}.
\]
The raw score vector is
\[
z=W_Uh_3+b.
\]
Substituting the numbers gives
\[
z
=
\begin{bmatrix}
2.0 & 1.0\\
1.0 & 0.5\\
0.5 & 1.0\\
-1.0 & 0.0
\end{bmatrix}
\begin{bmatrix}
1.0\\
0.5
\end{bmatrix}
+
\begin{bmatrix}
0.0\\
0.1\\
0.0\\
0.0
\end{bmatrix}
=
\begin{bmatrix}
2.5\\
1.35\\
1.0\\
-1.0
\end{bmatrix}.
\]
So the model assigns raw score \(2.5\) to \(\texttt{lookup\_order}\), raw score \(1.35\) to \(\texttt{escalate\_case}\), raw score \(1.0\) to \(\texttt{ask\_clarify}\), and raw score \(-1.0\) to \(\texttt{deny\_request}\). These are logits in the neural-network sense. They are not yet probabilities.

The numbers are artificial, but the structure is the same as in a real autoregressive language model with a much larger hidden dimension and vocabulary.

\subsubsection{Why the final-position vector can reflect earlier tokens}

The answer is attention. A transformer uses attention to let one token position combine information from other positions in the context. In an autoregressive language model, a position cannot use future tokens, but it can use earlier tokens. Therefore, the calculation that produces the final-position vector \(h_m\) can incorporate information from positions \(1,\ldots,m\), not just from the last token \(x_m\). This is why \(h_m\) can be used to score candidate next tokens based on the whole available prefix.

A single attention operation is often written in matrix form as
\[
\operatorname{Attention}(Q,K_{\mathrm{att}},V_{\mathrm{att}})
=
\softmax\left(\frac{QK_{\mathrm{att}}^\top}{\sqrt{d_{\mathrm{qk}}}}\right)V_{\mathrm{att}}.
\]
Here \(Q\) is the matrix of query vectors, \(K_{\mathrm{att}}\) is the matrix of key vectors, and \(V_{\mathrm{att}}\) is the matrix of value vectors. We use \(K_{\mathrm{att}}\) rather than \(K\) to avoid confusion with \(K\), the vocabulary size used elsewhere in the tutorial. The quantity \(d_{\mathrm{qk}}\) is the dimension of the query and key vectors; it is not the same object as \(d\), the hidden-state dimension used above.

The uppercase matrices stack the corresponding lowercase vectors. For a context with \(m\) token positions,
\[
Q=
\begin{bmatrix}
q_1^\top\\
q_2^\top\\
\vdots\\
q_m^\top
\end{bmatrix},
\qquad
K_{\mathrm{att}}=
\begin{bmatrix}
k_1^\top\\
k_2^\top\\
\vdots\\
k_m^\top
\end{bmatrix},
\qquad
V_{\mathrm{att}}=
\begin{bmatrix}
v_1^\top\\
v_2^\top\\
\vdots\\
v_m^\top
\end{bmatrix}.
\]
Thus, \(q_i\) is the query vector for position \(i\), \(k_i\) is the key vector for position \(i\), and \(v_i\) is the value vector for position \(i\).

For this tutorial, the important idea is simple: attention compares a query vector with key vectors, turns the resulting similarity scores into weights, and then uses those weights to average the value vectors. The query represents what a position is looking for. The keys represent what each position offers to be matched against the query. The values contain the information to be combined \citep{vaswani2017attention}.

Continuing the toy example above, suppose the current model context contains three tokens:
\[
x_1=\texttt{refund},\qquad
x_2=\texttt{order},\qquad
x_3=\texttt{please}.
\]
To keep the example small, consider only the attention calculation for the final position \(x_3\). Suppose the final position has query vector
\[
q_3=
\begin{bmatrix}
1\\
0
\end{bmatrix}.
\]
Suppose the three key vectors are
\[
k_1=
\begin{bmatrix}
0.5\\
0
\end{bmatrix},
\qquad
k_2=
\begin{bmatrix}
1.0\\
0
\end{bmatrix},
\qquad
k_3=
\begin{bmatrix}
0.2\\
0
\end{bmatrix}.
\]
The dot products between the final-position query \(q_3\) and the three keys are
\[
q_3^\top k_1=0.5,\qquad
q_3^\top k_2=1.0,\qquad
q_3^\top k_3=0.2.
\]
These numbers are similarity scores. In this toy example, the final position matches most strongly with \(x_2=\texttt{order}\), then with \(x_1=\texttt{refund}\), and least strongly with \(x_3=\texttt{please}\).

In this toy attention calculation, the query and key vectors have dimension \(d_{\mathrm{qk}}=2\). This is why the similarity scores are divided by \(\sqrt{2}\):
\[
\frac{1}{\sqrt{2}}(0.5,\;1.0,\;0.2)
\approx
(0.354,\;0.707,\;0.141).
\]
Applying softmax gives attention weights approximately equal to
\[
(0.309,\;0.440,\;0.251).
\]
These weights sum to one. They say that, in this attention operation, the final position places about 30.9 percent weight on the first position, 44.0 percent weight on the second position, and 25.1 percent weight on the third position.

Now suppose the corresponding value vectors are
\[
v_1=
\begin{bmatrix}
0.1\\
0.8
\end{bmatrix},
\qquad
v_2=
\begin{bmatrix}
0.9\\
0.2
\end{bmatrix},
\qquad
v_3=
\begin{bmatrix}
0.3\\
0.4
\end{bmatrix}.
\]
These value vectors are illustrative internal vectors for this one attention operation. They should not be confused with the rows of the hidden-state matrix \(H\) used above.

The attention output at the final position is the weighted average
\[
0.309v_1+0.440v_2+0.251v_3.
\]
Substituting the numbers,
\[
0.309
\begin{bmatrix}
0.1\\
0.8
\end{bmatrix}
+
0.440
\begin{bmatrix}
0.9\\
0.2
\end{bmatrix}
+
0.251
\begin{bmatrix}
0.3\\
0.4
\end{bmatrix}
\approx
\begin{bmatrix}
0.502\\
0.436
\end{bmatrix}.
\]
The exact numbers are not important. The important point is that the representation at the final position is not based only on the last token. It combines information associated with \(\texttt{refund}\), \(\texttt{order}\), and \(\texttt{please}\), with weights determined by the attention calculation.

This single attention output is an internal sub-step, not the final vector \(h_m\). In a real transformer, the final-position hidden vector emerges after many layers, attention heads, feed-forward transformations, residual connections, and normalization steps. The toy example omits those details. Its purpose is only to show why the final-position representation can incorporate information from earlier tokens in the context. This is what makes it reasonable for the model to use \(h_m\) to score candidate next tokens based on the whole available prefix rather than only the last token.

\subsubsection{Why the inference forward pass is deterministic under controlled conditions}

The attention example also illustrates a second point that matters for this tutorial. Once the token IDs, model weights, input shapes, and numerical execution conditions are fixed, the inference forward pass is a sequence of numerical operations: matrix products, scaling, softmax operations, weighted sums, normalizations, nonlinear activations, and output projections. In that controlled sense, the forward pass is deterministic.

This does not mean that all neural-network computation is always deterministic in every setting. During training, some techniques deliberately inject randomness. For example, dropout randomly zeroes some activations during training. At inference time, however, models are normally run in evaluation mode. PyTorch's Dropout documentation notes that, during evaluation, the dropout module computes the identity function \citep{pytorch-dropout}. Thus, standard inference disables this training-time source of randomness.

This distinction is central to the tutorial's argument. The model's forward pass can be deterministic under controlled conditions, while the decoder can still introduce stochasticity by sampling from the probability distribution over next tokens. In deployed systems, additional variation may also enter through numerical and serving details, context construction, orchestration, tools, and changing external environments.

\subsection{From logits to probabilities}

A probabilistic decoder needs a probability distribution. The raw scores are converted into probabilities using the softmax function.\footnote{The softmax function takes a vector of real numbers and returns positive numbers that sum to one. If \(z_i\) is larger than \(z_j\), then \(\exp(z_i)\) is larger than \(\exp(z_j)\), so larger scores receive larger probabilities. The denominator normalizes the values so that the total probability is one.}

For the current model context \(c_n\), let
\[
z(c_n)=[z_1(c_n),z_2(c_n),\ldots,z_K(c_n)]^\top
\]
be the column vector of raw scores assigned to the \(K\) vocabulary tokens. The temperature-scaled probability assigned to candidate token \(\tau_i\) is
\[
p_T(\tau_i\mid c_n)
=
\frac{\exp(z_i(c_n)/T)}
{\sum_{j=1}^K \exp(z_j(c_n)/T)}.
\]
In vector notation,
\[
p_T(\cdot\mid c_n)=\softmax(z(c_n)/T).
\]
The resulting vector
\[
p_T(\cdot\mid c_n)
=
[p_T(\tau_1\mid c_n),p_T(\tau_2\mid c_n),\ldots,p_T(\tau_K\mid c_n)]^\top
\]
is a probability distribution over the fixed vocabulary \(\Vocab\). When \(c_n\), \(T\), and the logit vector are clear from context, we may write this probability compactly as \(p_i\). This shorthand should not be read as saying that the probability assigned to \(\tau_i\) depends only on \(z_i\). Because of the denominator in the softmax, \(p_T(\tau_i\mid c_n)\) depends on the entire score vector \(z(c_n)\).

\paragraph{Continuing the toy numerical example.}
Return to the toy vocabulary
\[
\Vocab
=
\{\tau_1,\tau_2,\tau_3,\tau_4\}.
\]
For concreteness, let the four candidate next tokens be
\[
\begin{aligned}
\tau_1&=\texttt{lookup\_order}, &
\tau_2&=\texttt{escalate\_case},\\
\tau_3&=\texttt{ask\_clarify}, &
\tau_4&=\texttt{deny\_request}.
\end{aligned}
\]
From the previous subsection, suppose the raw score vector is
\[
z=[z_1,z_2,z_3,z_4]^\top=[2.5,\;1.35,\;1.0,\;-1.0]^\top.
\]
These raw scores are not probabilities. They do not sum to one, and one of them is negative. The softmax function converts them into positive numbers that sum to one.

At temperature \(T=1\), the unnormalized softmax weights are
\[
\exp(2.5)\approx 12.182,\qquad
\exp(1.35)\approx 3.857,\qquad
\exp(1.0)\approx 2.718,\qquad
\exp(-1.0)\approx 0.368.
\]
Their sum is
\[
12.182+3.857+2.718+0.368 \approx 19.125.
\]
Therefore,
\[
p_1
=
\frac{12.182}{19.125}
\approx 0.637,
\qquad
p_2
=
\frac{3.857}{19.125}
\approx 0.202,
\]
\[
p_3
=
\frac{2.718}{19.125}
\approx 0.142,
\qquad
p_4
=
\frac{0.368}{19.125}
\approx 0.019.
\]
Thus, at \(T=1\),
\[
p_1+p_2+p_3+p_4
\approx
0.637+0.202+0.142+0.019
=
1.000.
\]
The highest-probability next token is
\[
\tau_1=\texttt{lookup\_order}.
\]
A greedy decoder would select this token. A sampling decoder, however, may still select another candidate token because the other tokens receive positive probability.

The terminology is potentially confusing for readers familiar with discrete choice models. In discrete choice theory, the softmax form corresponds to multinomial logit choice probabilities \citep{train2009discretechoice}. In neural-network usage, however, the term ``logits'' usually refers to the raw pre-softmax score vector \(z\). Thus, the model first produces raw scores, commonly called logits in ML practice, and the softmax function converts those scores into probabilities. The terminology overlaps, but the objects are distinct.

\section{Decoding: turning probabilities into an output sequence}
\label{sec:decoding}

In this context, \emph{decoding} refers to the algorithmic process of selecting output tokens from the model's computed scores or probabilities. The model does not directly produce a complete paragraph in one step. It repeatedly computes a next-token distribution, the decoder selects a token, the selected token is appended to the context, and the process repeats.

\subsection{Deterministic decoding}

A deterministic decoder does not use random draws. The simplest case is greedy decoding. After the model has converted logits into probabilities, greedy decoding chooses
\[
o_n = \argmax_{t \in \Vocab} P_\theta(t \mid c_n),
\qquad
c_{n+1} = c_n \append o_n.
\]
The equation says that, among all vocabulary tokens \(t\in\Vocab\), the decoder selects the token assigned the highest probability by the foundation model given the current model context \(c_n\). The selected token \(o_n\) is appended to the context, producing \(c_{n+1}\). No pseudo-random number generator is used in this step. Hugging Face's generation documentation describes greedy search as selecting the most probable next token at each step \citep{hf-generation-strategies}.

Under fixed weights, fixed prompt, fixed numerical environment, and fixed decoding settings, greedy decoding is deterministic. This is why it is imprecise to say that a foundation model is inherently stochastic. The same model can often be run with deterministic decoding or with stochastic decoding. Under deterministic decoding, the output sequence need not be stochastic in the sampling sense. Determinism, however, is not the same thing as quality. Greedy decoding can be useful in settings where repeatability is valued, but it can also become repetitive because it repeatedly makes locally most likely next-token choices, a failure mode related to neural text degeneration \citep{holtzman2020curious}.

\subsection{Sampling-based decoding}

A sampling decoder uses the same probability distribution differently. Instead of choosing the highest-probability token, it draws from the next-token probability distribution. For this subsection, write
\[
p_i=p_T(\tau_i\mid c_n).
\]
If
\[
p_T(\cdot\mid c_n)=(p_1,p_2,\ldots,p_K),
\]
then the decoder samples a token
\[
o_n \sim \Categorical(p_1,p_2,\ldots,p_K).
\]
A categorical distribution is the distribution over one draw from a finite set of alternatives: token \(\tau_1\) is selected with probability \(p_1\), token \(\tau_2\) with probability \(p_2\), and so on.

In a computer implementation, this sampling is typically done with a pseudo-random number generator, not physical randomness.\footnote{A pseudo-random number generator is a deterministic algorithm that produces a sequence of numbers designed to behave like random draws for practical purposes. The initial internal state is often controlled by a seed. Same seed plus same sequence of consumed draws gives the same pseudo-random stream.} A simple way to see the sampling step is to form cumulative probabilities
\[
F_i=\sum_{j=1}^i p_j.
\]
The decoder draws a pseudo-random number \(u_n\in[0,1)\) and selects the smallest index \(i\) such that
\[
F_i > u_n.
\]
For example, if the probabilities over four tokens are
\[
(0.10,0.20,0.30,0.40),
\]
then the unit interval can be divided into intervals of lengths \(0.10\), \(0.20\), \(0.30\), and \(0.40\). A pseudo-random number falling in a given interval selects the corresponding token. PyTorch's \texttt{torch.\allowbreak multinomial} is one implementation of this kind of sampling and accepts an explicit \texttt{generator} argument \citep{pytorch-multinomial}. A self-contained sampler that uses no AI model is given in Appendix~\ref{app:toy-sampler}. Figure~\ref{fig:inverse-cdf} illustrates this lookup for the example above.

\begin{figure}[H]
\centering
\begin{tikzpicture}[x=1mm,y=1mm]
  \def\H{11}
  \fill[black!10] (0,0)  rectangle (12,\H);
  \fill[black!16] (12,0) rectangle (36,\H);
  \fill[stochamber!35] (36,0) rectangle (72,\H);
  \fill[black!13] (72,0) rectangle (120,\H);
  \draw[black!55,line width=0.5pt] (0,0) rectangle (120,\H);
  \foreach \x in {12,36,72} {\draw[black!55,line width=0.4pt] (\x,0) -- (\x,\H);}

  \node[font=\scriptsize,align=center] at (6,\H/2)  {$\tau_1$\\$0.10$};
  \node[font=\scriptsize,align=center] at (24,\H/2) {$\tau_2$\\$0.20$};
  \node[font=\scriptsize,align=center,stochamber!85!black] at (46,\H/2) {$\tau_3$\\$0.30$};
  \node[font=\scriptsize,align=center] at (96,\H/2) {$\tau_4$\\$0.40$};

  \foreach \x/\lab in {0/0.0, 12/0.10, 36/0.30, 72/0.60, 120/1.0}
    {\draw[black!50] (\x,0) -- (\x,-2);
     \node[font=\scriptsize,black!60,below] at (\x,-2) {\lab};}

  \draw[sysred,line width=0.8pt] (54,-4) -- (54,\H+5);
  \fill[sysred] (54,\H+5) circle (0.9);
  \node[font=\small,sysred,above] at (54,\H+5.5) {$u=0.45$};
  \node[font=\scriptsize,align=center,black!70,below=7mm] at (54,-4)
        {$u$ lands in $\tau_3$'s interval $[0.30,0.60)$, so token $\tau_3$ is selected};
\end{tikzpicture}
\caption{Sampling as a lookup on the unit interval. The probabilities $(0.10,0.20,0.30,0.40)$ partition $[0,1)$ into four sub-intervals. A single pseudo-random draw $u$ selects the token whose interval contains it. Reusing the same $u$ with the same intervals always selects the same token, which is why sampling is reproducible when the random stream is controlled.}
\label{fig:inverse-cdf}
\end{figure}
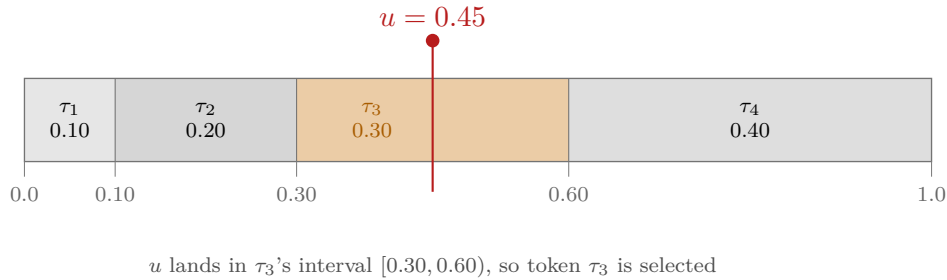

This is the clearest sense in which model output is stochastic during ordinary sampling-based generation: the model computes a distribution, and the decoder draws from that distribution.

\subsection{Temperature, top-\texorpdfstring{$k$}{k}, and top-\texorpdfstring{$p$}{p}}

Temperature changes the shape of the distribution before sampling:
\[
p_T(\tau_i\mid c_n)=\frac{\exp(z_i(c_n)/T)}{\sum_{j=1}^K \exp(z_j(c_n)/T)}.
\]
Higher \(T\) flattens the distribution, giving lower-ranked tokens more probability mass and thereby increasing the role of randomness during sampling. Conversely, lower \(T\) concentrates mass on the highest-scoring tokens, making the sampled choice more nearly deterministic.
As \(T\) approaches zero from above, the distribution concentrates on the highest-scoring token, and the procedure approaches greedy decoding.\footnote{The formula itself is defined for \(T>0\). Setting \(T=0\) literally would require division by zero. Implementations that expose \texttt{temperature = 0} usually special-case it as greedy or near-greedy decoding rather than substituting zero into the formula. A related subtlety: if the maximizing logit is unique, the \(T\to 0^+\) limit places all mass on one token, but if several tokens tie for the maximum, that limit is uniform over the tied set, whereas algorithmic greedy decoding, for example \texttt{argmax}, breaks the tie deterministically, typically by lowest index, so practical greedy decoding stays deterministic.}

Using the same toy raw scores \(z=[2.5,1.35,1.0,-1.0]^\top\), the temperature-scaled probabilities are approximately:

\begin{center}
\footnotesize
\begin{tabular}{ccccc}
\toprule
\(T\)
&
\(p_T(\tau_1\mid c_n)\)
&
\(p_T(\tau_2\mid c_n)\)
&
\(p_T(\tau_3\mid c_n)\)
&
\(p_T(\tau_4\mid c_n)\)
\\
\midrule
0.5 & 0.869 & 0.087 & 0.043 & 0.001\\
1.0 & 0.637 & 0.202 & 0.142 & 0.019\\
2.0 & 0.453 & 0.255 & 0.214 & 0.079\\
\bottomrule
\end{tabular}
\end{center}

Entries are rounded to three decimals, so rows may not sum exactly to \(1.000\). Figure~\ref{fig:temperature} visualizes the same calculation. The raw scores are unchanged across panels. Only the temperature changes.

\begin{figure}[H]
\centering
\begin{tikzpicture}[x=1mm,y=1mm]
  \def\barw{5}
  \def\step{8}

  \begin{scope}[xshift=0mm]
    \draw[black!45,line width=0.4pt] (-2,0) -- (36,0);
    \fill[agentblue!55] (0,0) rectangle (5,26.1);
    \fill[agentblue!55] (8,0) rectangle (13,2.6);
    \fill[agentblue!55] (16,0) rectangle (21,1.3);
    \fill[agentblue!55] (24,0) rectangle (29,0.1);
    \node[font=\small\bfseries,black!75] at (15,34) {$T=0.5$};
    \node[font=\scriptsize,black!60] at (15,30) {sharp};
    \node[font=\scriptsize,rotate=45,anchor=east] at (2.5,-2) {$\tau_1$};
    \node[font=\scriptsize,rotate=45,anchor=east] at (10.5,-2) {$\tau_2$};
    \node[font=\scriptsize,rotate=45,anchor=east] at (18.5,-2) {$\tau_3$};
    \node[font=\scriptsize,rotate=45,anchor=east] at (26.5,-2) {$\tau_4$};
    \node[font=\scriptsize,black!70] at (2.5,28.5) {.869};
    \node[font=\scriptsize,black!70] at (10.5,5.0) {.087};
    \node[font=\scriptsize,black!70] at (18.5,3.7) {.043};
    \node[font=\scriptsize,black!70] at (26.5,2.4) {.001};
  \end{scope}

  \begin{scope}[xshift=54mm]
    \draw[black!45,line width=0.4pt] (-2,0) -- (36,0);
    \fill[agentblue!55] (0,0) rectangle (5,19.1);
    \fill[agentblue!55] (8,0) rectangle (13,6.1);
    \fill[agentblue!55] (16,0) rectangle (21,4.3);
    \fill[agentblue!55] (24,0) rectangle (29,0.6);
    \node[font=\small\bfseries,black!75] at (15,34) {$T=1.0$};
    \node[font=\scriptsize,black!60] at (15,30) {baseline};
    \node[font=\scriptsize,rotate=45,anchor=east] at (2.5,-2) {$\tau_1$};
    \node[font=\scriptsize,rotate=45,anchor=east] at (10.5,-2) {$\tau_2$};
    \node[font=\scriptsize,rotate=45,anchor=east] at (18.5,-2) {$\tau_3$};
    \node[font=\scriptsize,rotate=45,anchor=east] at (26.5,-2) {$\tau_4$};
    \node[font=\scriptsize,black!70] at (2.5,21.6) {.637};
    \node[font=\scriptsize,black!70] at (10.5,8.5) {.202};
    \node[font=\scriptsize,black!70] at (18.5,6.7) {.142};
    \node[font=\scriptsize,black!70] at (26.5,3.0) {.019};
  \end{scope}

  \begin{scope}[xshift=108mm]
    \draw[black!45,line width=0.4pt] (-2,0) -- (36,0);
    \fill[agentblue!55] (0,0) rectangle (5,13.6);
    \fill[agentblue!55] (8,0) rectangle (13,7.6);
    \fill[agentblue!55] (16,0) rectangle (21,6.4);
    \fill[agentblue!55] (24,0) rectangle (29,2.4);
    \node[font=\small\bfseries,black!75] at (15,34) {$T=2.0$};
    \node[font=\scriptsize,black!60] at (15,30) {flatter};
    \node[font=\scriptsize,rotate=45,anchor=east] at (2.5,-2) {$\tau_1$};
    \node[font=\scriptsize,rotate=45,anchor=east] at (10.5,-2) {$\tau_2$};
    \node[font=\scriptsize,rotate=45,anchor=east] at (18.5,-2) {$\tau_3$};
    \node[font=\scriptsize,rotate=45,anchor=east] at (26.5,-2) {$\tau_4$};
    \node[font=\scriptsize,black!70] at (2.5,16.0) {.453};
    \node[font=\scriptsize,black!70] at (10.5,10.0) {.255};
    \node[font=\scriptsize,black!70] at (18.5,8.8) {.214};
    \node[font=\scriptsize,black!70] at (26.5,4.8) {.079};
  \end{scope}

  \node[font=\footnotesize,black!70] at (70,-14)
    {Same raw scores \(z=[2.5,1.35,1.0,-1.0]^\top\); only the temperature \(T\) changes.};
\end{tikzpicture}
\caption{Temperature reshapes the toy next-token distribution. Lower temperature concentrates probability on the highest-scoring token, while higher temperature spreads more probability mass to lower-ranked tokens.}
\label{fig:temperature}
\end{figure}

Top-\(k\) and top-\(p\) sampling restrict the distribution before sampling. In top-\(k\) sampling, let \(\Vocabk(c_n)\subseteq \Vocab\) be the set of the \(k\) highest-probability tokens under the current model context. For readability, write \(p_i=p_T(\tau_i\mid c_n)\). The restricted distribution is
\[
\tilde p_i
=
\begin{cases}
\dfrac{p_i}{\sum_{\ell:\,\tau_\ell\in \Vocabk(c_n)}p_\ell}, & \tau_i\in \Vocabk(c_n),\\[9pt]
0, & \tau_i\notin\Vocabk(c_n).
\end{cases}
\]
The equation says: keep only the \(k\) most likely tokens, set all other probabilities to zero, and rescale the remaining probabilities so they again sum to one. The decoder then samples from \(\tilde p\).
In top-\(p\), or nucleus, sampling, sort the tokens from most probable to least probable. Let \(\Vocabp(c_n)\subseteq\Vocab\) be the smallest prefix of the sorted tokens whose cumulative probability is at least \(p\). The decoder sets the probability of all other tokens to zero, renormalizes over \(\Vocabp(c_n)\), and samples from the resulting distribution. Nucleus sampling was proposed as a way to avoid the degeneracy of deterministic decoding while also avoiding the long tail of low-probability tokens \citep{holtzman2020curious}.

In temperature sampling, top-\(k\) sampling, and top-\(p\) sampling, the distribution is modified before token selection. If the final selection is by sampling, the pseudo-RNG enters at that point.

\subsection{Why text generation is sequential}

After the decoder selects a token, the token is appended to the model context. If the current model context is \(c_n\) and the selected token is \(o_n\), then
\[
c_{n+1}=c_n\append o_n.
\]
This gives the autoregressive loop:
\[
\text{score tokens}\rightarrow \text{decode one token}\rightarrow \text{append token}\rightarrow \text{repeat}.
\]
Token \(n+1\) cannot be generated in the ordinary autoregressive way until token \(n\) has been selected, because the selected token becomes part of the next model context.

This sequential dependence is one reason text generation can be slow. Many computations inside a single forward pass can be parallelized efficiently on GPUs. Contemporary implementations also use a key-value (KV) cache, which stores attention keys and values from previous tokens so the model need not recompute all earlier token representations from scratch.\footnote{A KV cache stores the ``key'' and ``value'' matrices used by attention for previous tokens. It speeds up generation but does not remove the sequential dependency that the next accepted token depends on the previous accepted token.} But the high-level generation loop remains sequential at the token level. Long answers usually take longer because they require more decode steps. Speculative decoding can accelerate this loop by proposing several candidate tokens with a smaller model and then verifying them with the larger model, but the final accepted sequence still has to be consistent with the autoregressive model.\footnote{Speculative decoding is a speed technique. It should not be confused with a different reasoning algorithm. It proposes candidate future tokens, verifies them, and accepts or rejects them under rules intended to preserve the target model's distribution.}

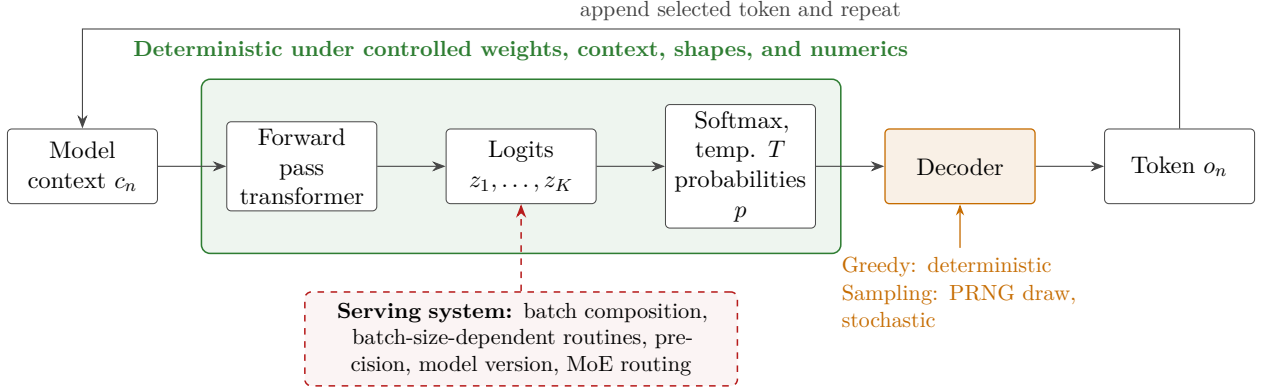
\begin{figure}[t]
\centering
\resizebox{\textwidth}{!}{%
\begin{tikzpicture}[node distance=11mm]
  \tikzset{
    box/.style    = {draw=black!72, rounded corners=2pt, fill=white,
                     minimum width=23mm, minimum height=12mm, align=center,
                     font=\small, text width=22mm, inner sep=3pt},
    sampler/.style= {box, draw=stochamber, line width=0.6pt, fill=stochamber!10},
    sys/.style    = {draw=sysred, dashed, line width=0.7pt, rounded corners=3pt,
                     fill=sysred!5, align=center, font=\footnotesize,
                     text width=66mm, inner sep=5pt},
    flow/.style   = {-{Stealth[length=2.2mm]}, black!72, line width=0.5pt},
    loop/.style   = {-{Stealth[length=2.2mm]}, black!72, line width=0.5pt},
    perturb/.style= {-{Stealth[length=2.2mm]}, sysred, dashed, line width=0.7pt}
  }

  \node[box] (ctx)                       {Model context $c_n$};
  \node[box, right=of ctx] (fwd)         {Forward pass\\transformer};
  \node[box, right=of fwd] (logit)       {Logits\\$z_1,\dots,z_K$};
  \node[box, right=of logit] (soft)      {Softmax, temp. $T$\\probabilities $p$};
  \node[sampler, right=of soft] (samp)   {Decoder};
  \node[box, right=of samp] (tok)        {Token $o_n$};

  \draw[flow] (ctx)   -- (fwd);
  \draw[flow] (fwd)   -- (logit);
  \draw[flow] (logit) -- (soft);
  \draw[flow] (soft)  -- (samp);
  \draw[flow] (samp)  -- (tok);

  \begin{scope}[on background layer]
    \node[draw=detgreen, line width=0.7pt, rounded corners=4pt,
          fill=detgreen!8, fit=(fwd)(logit)(soft),
          inner sep=4mm] (core) {};
  \end{scope}
  \node[detgreen, font=\footnotesize\bfseries, above=1.5mm of core.north]
        {Deterministic under controlled weights, context, shapes, and numerics};

  \draw[loop] (tok.north) -- ++(0,16mm) -| (ctx.north)
        node[pos=0.2, above, font=\footnotesize, black!75]
        {append selected token and repeat};

  \node[stochamber, font=\footnotesize, align=left, below=7mm of samp,
        text width=38mm] (sampnote)
        {Greedy: deterministic\\[1pt]
         Sampling: PRNG draw, stochastic};
  \draw[stochamber, line width=0.6pt, -{Stealth[length=1.8mm]}]
        (sampnote.north) -- (samp.south);

  \node[sys, below=14mm of logit] (sys)
        {\textbf{Serving system:} batch composition, batch-size-dependent routines, precision, model version, MoE routing};
  \draw[perturb] (sys.north) -- (logit.south);
\end{tikzpicture}}
\caption{The token-generation pipeline. The model's forward pass computes logits and probabilities. The explicit source of sampling stochasticity is the decoder. The serving layer can also perturb logits, so deterministic decoding does not guarantee identical outputs in every deployed system.}
\label{fig:token-pipeline}
\end{figure}

Figure~\ref{fig:token-pipeline} summarizes a single decoding step end to end.

\section{From token choices to agent trajectories}
\label{sec:amplification}

Because the output of one step becomes the input for the next, variation can compound rapidly. A single model call generates a sequence of tokens, which the agent may subsequently parse into actionable formats such as text, JSON, code, a function call, a tool argument, or a structured plan. At agent round \(r\), write the generated sequence as
\[
Y_r=(o_{r,1},o_{r,2},\ldots,o_{r,L_r}).
\]
If sampling changes \(o_{r,5}\), the remaining tokens may change because the subsequent context changes. At the token level,
\[
c_{r,6}=c_{r,5}\append o_{r,5}.
\]
At the model-output level, a changed token sequence may produce a different generated sentence, JSON object, code fragment, or plan step. At the agent-action level, a changed output sequence can change the parsed action:
\[
A_r=\Parse(Y_r).
\]
A different action can produce a different observation \(O_r\), and a different observation changes the next state \(S_{r+1}\). At the workflow level, the same chain may lead to a different support routing decision, refund recommendation, code patch, or compliance consequence. Thus, variation compounds across two nested loops:

\begin{enumerate}[leftmargin=*]
\item \textbf{The token loop:} score next tokens, decode one token, append it, and repeat.
\item \textbf{The agent loop:} generate a model output, parse it as an action, observe the environment, update state, and repeat.
\end{enumerate}

Because of these nested loops, agentic systems can exhibit more consequential variability than ordinary chat completions. In ordinary chat, a different sampled phrase may simply be a different wording. In an agent, a different sampled token can alter a tool call, a database query, a code patch, or a decision to continue or stop.

\begin{figure}[H]
\centering
\resizebox{\textwidth}{!}{%
\begin{tikzpicture}[node distance=6mm]
  \tikzset{
    tok/.style   = {draw=black!60, fill=white, rounded corners=2pt, font=\footnotesize,
                    minimum height=8mm, text width=8mm, align=center, line width=0.5pt},
    fork/.style  = {diamond, draw=stochamber, fill=stochamber!12, aspect=1.7,
                    inner sep=1pt, font=\footnotesize, align=center, line width=0.7pt},
    act/.style   = {draw=toolpurple, fill=toolpurple!8, rounded corners=2pt, font=\footnotesize,
                    align=center, text width=30mm, minimum height=9mm, line width=0.6pt},
    outcome/.style   = {draw=sysred, fill=sysred!7, rounded corners=2pt, font=\footnotesize,
                    align=center, text width=30mm, minimum height=9mm, line width=0.6pt},
    flow/.style  = {-{Stealth[length=2mm]}, black!70, line width=0.5pt},
    branch/.style= {-{Stealth[length=2mm]}, stochamber, line width=0.8pt}
  }
  \node[tok] (o1) {$o_1$};
  \node[tok, right=of o1] (o2) {$o_2$};
  \node[tok, right=of o2] (o3) {$o_3$};
  \node[tok, right=of o3] (o4) {$o_4$};
  \draw[flow] (o1)--(o2); \draw[flow] (o2)--(o3); \draw[flow] (o3)--(o4);

  \node[fork, right=10mm of o4] (fork) {sample\\$o_5$};
  \draw[flow] (o4)--(fork);

  \node[act, above right=4mm and 12mm of fork] (au) {$o_5=$ \texttt{lookup\_order}};
  \node[act, below right=4mm and 12mm of fork] (ad) {$o_5=$ \texttt{escalate\_case}};
  \draw[branch] (fork) -- (au);
  \draw[branch] (fork) -- (ad);

  \node[outcome, right=10mm of au] (ou) {read order,\\answer customer};
  \node[outcome, right=10mm of ad] (od) {open ticket,\\human handoff};
  \draw[flow] (au)--(ou);
  \draw[flow] (ad)--(od);

  \begin{scope}[on background layer]
    \node[draw=black!45, dashed, rounded corners=4pt, fill=black!3,
          fit=(o1)(o4), inner sep=3mm,
          label={[font=\scriptsize,black!60]below:identical prefix (same tokens in both runs)}] {};
  \end{scope}
  \node[font=\scriptsize, stochamber, align=center, above=1mm of fork, text width=30mm]
        {one different sampled token};
\end{tikzpicture}}
\caption{How a single token flips a trajectory. Two runs share an identical token prefix, then sampling selects a different token at one step. Parsed as a tool call, that one difference sends the agent down a different branch, with a different action and a materially different outcome. This is the amplification from token-level sampling to agent-level behavior.}
\label{fig:amplification}
\end{figure}
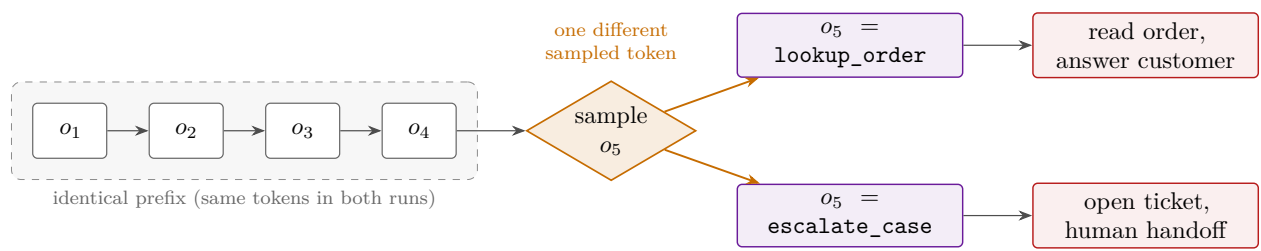

Consider the running customer-support example. The model might generate one of the following tool calls as JSON:

\begin{lstlisting}[basicstyle=\ttfamily\small,frame=single,breaklines=true]
{"tool": "lookup_order", "arguments": {"order_id": "A173"}}
{"tool": "escalate_case", "arguments": {"reason": "refund dispute"}}
\end{lstlisting}

A few different tokens can change the business action. Similarly, a data agent may generate
\begin{lstlisting}[basicstyle=\ttfamily\small,frame=single,breaklines=true]
WHERE status = 'active'
\end{lstlisting}
versus
\begin{lstlisting}[basicstyle=\ttfamily\small,frame=single,breaklines=true]
WHERE status IN ('active', 'pending')
\end{lstlisting}
The difference is small as text but material as a database query.
Figure~\ref{fig:amplification} traces how one differing token sends the agent down a different branch.\footnote{For readability the figure labels a branch by a whole identifier such as \texttt{lookup\_order}. A real tokenizer splits such identifiers into subword tokens, so the branch point is the first subword token that differs between runs. The amplification argument is unchanged: once any token in the serialized action differs, the parsed action and its consequences can diverge.}

\section{What stochasticity does not mean}
\label{sec:not-mean}

\subsection{It does not mean the network explores random reasoning paths}

The terminology of ``reasoning paths'' is a common source of confusion. A reasoning path is a candidate chain of intermediate steps, partial conclusions, or candidate continuations leading from a prompt to an answer. A hypothetical search system might try one path that solves a problem algebraically, another path that uses a numerical example, and another path that looks for a counterexample. It might evaluate those paths and choose among them.

Some AI applications do implement explicit search-like or multi-candidate procedures. The ReAct framework studies interleaved reasoning traces and actions in language-model agents \citep{yao2022react}. Toolformer studies models that decide which APIs to call, when to call them, and what arguments to pass \citep{schick2023toolformer}. Multi-agent systems such as AutoGen explore designs in which multiple LLM-based agents converse and coordinate \citep{wu2023autogen}. These are agent-level or application-level designs around model calls.

A standard dense transformer forward pass, however, does not internally search or randomly branch. Given the same token IDs, weights, and controlled numerical execution, it applies the same layers in the same order: attention operations, matrix multiplications, feed-forward layers, normalization steps, and output projection. The usual stochasticity in generation enters when the decoder samples output tokens.

This point also applies to many ``reasoning'' or extended-thinking models as users experience them. A visible or hidden reasoning trace is still generated as a token sequence, possibly with different prompting, training, or inference scaffolding. Variation across traces does not by itself imply that the neural network used a hidden random path selector. It may simply reflect sampling over a longer chain of intermediate tokens. A vendor or application may add search, self-consistency, reranking, verifier models, tool execution, or other inference-time procedures around the model. Such procedures can explore multiple candidate trajectories, but they are additional application-level or inference-time mechanisms rather than evidence that a single standard dense transformer forward pass randomly branches internally.

Mixture-of-experts models require a small clarification.\footnote{A mixture-of-experts model routes each token to a subset of specialized sub-networks, often called experts, rather than applying the same feed-forward sub-network to every token.} MoE routing is a kind of data-dependent routing decision inside the model. It is not necessarily random. Given the same input, same weights, same batch composition, and same routing implementation, it can be deterministic. It is therefore not a counterexample to the main point. It is a routing decision, not a coin flip. In deployed systems, however, MoE routing can interact with batch composition and load-balancing details, so it belongs among practical sources of variation.

\subsection{It does not mean irreproducible in principle}

Stochastic sampling can be exactly reproducible under matched conditions. Suppose two model replicas \(i\) and \(j\) have the same weights, the same tokenizer, the same decoding code, the same model context, a deterministic forward pass, and synchronized pseudo-RNG streams. At token step \(n\), assume
\[
c_{i,n}=c_{j,n}, \qquad R_{i,n}=R_{j,n}.
\]
The controlled forward pass gives the same probability distribution:
\[
p_{i,n}=p_{j,n}.
\]
The same RNG state gives the same pseudo-random draw:
\[
u_{i,n}=u_{j,n}.
\]
The same sampler applied to the same probability distribution and the same draw gives
\[
o_{i,n}=o_{j,n}.
\]
After appending the same token, the model contexts remain equal:
\[
c_{i,n+1}=c_{j,n+1}.
\]
The RNG states also advance in the same way. By induction, the two replicas generate the same token sequence.

The condition ``controlled forward pass'' matters. This induction argument assumes that equal contexts produce equal logits. That is true in the toy central processing unit (CPU) demonstration in Appendix~\ref{app:twin-demo}, where we use evaluation mode, fixed weights, fixed shapes, manual one-token sampling, and explicit pseudo-RNG generators. It may fail in a deployed service if the numerical or serving conditions perturb logits. The next subsection explains that distinction.

This is why ``stochastic'' should not be equated with ``unrepeatable''. A pseudo-random generator can make a sampled process reproducible when the seed, model, software, hardware, inputs, and sequence of random draws are controlled. Some commercial APIs expose seed-like controls, but they may still describe determinism as best-effort because backend infrastructure can change \citep{openai-seed-cookbook}.

\subsection{It does not necessarily mean the GPU used a pseudo-RNG}

A common explanation for irreproducibility at temperature zero is the finite precision of GPU arithmetic. While grounded in a true mathematical property, namely that floating-point arithmetic is finite precision and not perfectly associative, this explanation remains incomplete for LLM inference. In exact real arithmetic,
\[
(a+b)+c=a+(b+c).
\]
In floating-point arithmetic, the two sides can differ slightly because intermediate results are rounded.

But the simple ``concurrency plus floating-point'' explanation is not the whole story for LLM inference. On GPUs, operations such as matrix multiplication, normalization, or attention are carried out by low-level GPU routines called kernels. He and collaborators at Thinking Machines Lab argue that, for common LLM inference forward passes, individual kernels can be deterministic for fixed input shapes, while user-visible nondeterminism often arises because inference servers are not \emph{batch-invariant} \citep{he2025nondeterminism}. Batch invariance means that the output for one request should be the same regardless of what other requests share the same batch. In many practical serving systems, your prompt is silently batched with other users' prompts. If the batch size or shape changes, the library may use a different accumulation pattern, tiling strategy, or kernel path. Continuous or dynamic batching makes this issue especially intuitive: requests can be batched, unbatched, and rebatched as tokens are generated. The numerical path for a request may therefore depend not only on its own prompt, but also on which other requests share the server at particular token steps. The forward pass can be deterministic for a given batch while not being invariant to which batch your request lands in.

This distinction is important. The GPU need not be using a pseudo-RNG. The serving system may simply process your request in different batch contexts across runs. The model call can then receive slightly different logits, even under greedy decoding. If two tokens are nearly tied, a tiny logit perturbation can flip the highest-probability token. Once one token differs, the subsequent context differs, and the rest of the sequence can diverge.

PyTorch's reproducibility documentation makes a related distinction. Seeding controls random number generators, such as PyTorch, Python, NumPy, and DataLoader worker RNGs. Separately, settings such as \texttt{torch.use\_deterministic\_algorithms(True)} configure PyTorch to use deterministic implementations where available, or to raise an error when an operation is known to be nondeterministic without a deterministic alternative \citep{pytorch-reproducibility}. Seeding a PRNG addresses sampling randomness; deterministic kernels and batch-invariant serving address numerical and systems-level reproducibility.

\subsection{It does not mean deterministic decoding is always safe}

Greedy decoding removes the explicit sampling draw. It does not remove every source of variation in an agentic system: the environment can change, tools can fail, databases can update, guardrails can route cases differently, and a service provider can update a model or change a backend.

Therefore, ``temperature zero'' should be read narrowly. It usually means that the decoder is no longer intentionally sampling from a distribution. It does not guarantee that a deployed agentic application will reliably trace the same trajectory.

\section{Implications for tracing and governing agentic variability}
\label{sec:tracing-governance}

This final section translates the preceding analysis into diagnostic and governance principles for tracing agentic variability. The focus is not benchmark evaluation alone, but identifying where variation entered, whether it mattered, and whether it can be controlled, replayed, or only measured distributionally.

For a manager, product owner, instructor, researcher, or technical evaluator, the key question is not simply, ``Is the model stochastic?'' The better question is, ``At which layer can variation enter, and what variation matters for this use case?'' The following principles follow from the typology and the preceding technical analysis.

\begin{enumerate}[leftmargin=*]
\item \textbf{Distinguish sources, manifestations, and consequences of variation.} A sampled token is a source of variation at the decoding layer. A different generated JSON object, plan sentence, or code fragment is a manifestation at the model-output layer. A different tool call, database query, file inspection, or code edit is a manifestation at the agent-action layer. A different refund decision, customer routing, compliance exposure, or pull request outcome is a business-process consequence.

\item \textbf{Distinguish surface-form variation from action-level variation.} Two responses may differ in wording while remaining equivalent for the task. By contrast, two tool calls, database queries, code edits, or escalation decisions may differ in ways that change cost, risk, compliance exposure, or user experience.

\item \textbf{Trace complete agent trajectories, not only final outputs.} For agentic systems, logs should include constructed context, model output, parsed action, tool call, tool result, validation outcome, retry path, guardrail decision, and state update. The final answer alone often hides where variation entered.

\item \textbf{Use deterministic or seeded runs as diagnostic controls.} Deterministic decoding and fixed pseudo-random seeds are useful for pre-production regression testing, classroom demonstrations, and post-incident analysis. In pre-production testing, they help isolate whether a change in behavior is caused by prompt assembly, tool schemas, parser logic, application state, or model version. In post-incident analysis, they can help replay or approximate a problematic trajectory if the original context, tool observations, decoding settings, and relevant metadata were logged. These controls do not imply that deployed agents should always use deterministic decoding, nor do they show that production behavior will be deterministic.

\item \textbf{Freeze or replay the environment when testing reproducibility.} If the agent uses a live database, search index, file system, clock, external API, or human response, then a repeated run is not the same experiment unless those inputs are fixed, mocked, snapshotted, or replayed.

\item \textbf{Measure production behavior distributionally when variation is expected.} If the deployed agent uses sampling, retrieval, live tools, retries, or changing environments, a single run is often insufficient. Relevant evidence includes success rates, divergence points, failure modes, variance across runs, sensitivity to prompts, and sensitivity to tool outputs.

\item \textbf{Govern consequential actions, not only generated text.} In agentic systems, the consequential output may be a tool call, refund action, code patch, database update, email, or escalation. Guardrails should validate actions and arguments before execution, especially when actions affect customers, records, money, security, or compliance.
\end{enumerate}

A useful summary is:

\begin{quote}
A foundation model can compute next-token probabilities deterministically, a decoder can sample from those probabilities stochastically, and an agentic system can amplify sampled token differences into different plans, tool calls, observations, and outcomes. In deployed agents, changing environments and serving systems add further sources of variation.
\end{quote}

The appendices provide reproducible demonstrations for readers who want to see the sampling and seeding claims operationalized in code. They are not required for following the main conceptual argument, but they make the reproducibility claims concrete.

\appendix

\section{Notation}
\label{app:notation}

{\footnotesize
\setlength{\tabcolsep}{4pt}%
\renewcommand{\arraystretch}{1.05}%
\begin{longtable}{P{0.18\textwidth}P{0.72\textwidth}}
\caption{Notation used throughout the tutorial.}\label{tab:notation}\\
\toprule
\textbf{Symbol} & \textbf{Meaning} \\
\midrule
\endfirsthead
\toprule
\textbf{Symbol} & \textbf{Meaning} \\
\midrule
\endhead
\(r\) & Agent round. One round may include one model call, one parsed action, one observation, and a state update. \\
\(S_r\) & Agent state before round \(r\). \\
\(G\) & User's goal or task. \\
\(C_r\) & Conversation history available to the agent at round \(r\). \\
\(M_r\) & Memory, retrieved material, or other stored information. \\
\(\mathsf{Tools}_r\) & Tools available to the agent at round \(r\). \\
\(E_r\) & Current environment state at round \(r\), including relevant live data, files, database contents, API status, web observations, user interactions, or time. \\
\(Y_r\) & Output sequence generated by the model call at agent round \(r\). \\
\(L_r\) & Length of the output sequence \(Y_r\) generated at round \(r\). \\
\(L\) & Number of future tokens in a hypothetical continuation, as in the \(K^L\) counting argument. \\
\(A_r\) & Action parsed from \(Y_r\), such as a text response, tool call, code edit, or database query. \\
\(O_r\) & Observation returned by a tool or environment after action \(A_r\). \\
\(\mathcal{E}\) & Environment/tool response map that executes an action and returns an observation, possibly depending on the current environment state. \\
\(\Vocab\) & The model's fixed token vocabulary. \\
\(K=|\Vocab|\) & Vocabulary size. \\
\(\tau_i\) & The \(i\)th candidate token in the vocabulary. \\
\(m\) & Current model-context length, that is, the number of tokens in the serialized input sequence currently being processed by the model. \\
\(n\) & Token-generation step or token-position index. \\
\(c_n\) & Model context available before producing generated token \(o_n\). \\
\(o_n\) & Token produced at generation step \(n\); appending it yields \(c_{n+1}\). \\
\(d\) & Hidden dimension of the model; each token position is represented by a vector in \(\R^d\). \\
\(d_{\mathrm{qk}}\) & Query/key dimension in an attention operation, used in the scaling term \(\sqrt{d_{\mathrm{qk}}}\). \\
\(H\) & Matrix of hidden representations, \(H\in\R^{m\times d}\), with one row per token position and one column per hidden feature. \\
\(H_{i,:}\) & Row \(i\) of \(H\), with all columns included. The colon means all columns. \\
\(h_m\) & Hidden vector at the final position of a model context of length \(m\), used to predict the next token. \\
\(W_U\) & Output projection, or unembedding, matrix that maps the final hidden vector to one score per vocabulary token. \\
\(b\) & Bias vector in the output projection. \\
\(\theta\) & The model's parameters, or weights; the subscripts in \(f_\theta\) and \(P_\theta\) indicate dependence on them. \\
\(z_i(c_n)\) & Raw unnormalized score, or neural-network logit, assigned to token \(\tau_i\) when the current model context is \(c_n\). \\
\(z(c_n)\) & Column vector of raw scores \([z_1(c_n),\ldots,z_K(c_n)]^\top\) for the current model context. \\
\(p_T(\tau_i\mid c_n)\) & Probability assigned to token \(\tau_i\), given model context \(c_n\), after applying softmax with temperature \(T\). \\
\(p_i\) & Compact shorthand for \(p_T(\tau_i\mid c_n)\) when \(T\), \(c_n\), and the logit vector are clear. \\
\(P_\theta(\cdot\mid c_n)\) & The model's native next-token distribution given model context \(c_n\), typically corresponding to temperature \(T=1\) before decoder-specific reshaping. \\
\(T\) & Temperature parameter. Higher values flatten the distribution; lower values sharpen it. \\
\(R_r\) & Pseudo-random number generator state used in model call \(r\). \\
\(u_n\) & A pseudo-random draw used to sample a token at token step \(n\). \\
\(F_i\) & Cumulative probability \(F_i=\sum_{j=1}^{i} p_j\) of the first \(i\) tokens; the sampling lookup selects the smallest \(i\) with \(F_i > u_n\). \\
\(D_\lambda\) & Decoding procedure with parameters \(\lambda\), such as temperature, top-\(k\), or top-\(p\). \\
\(\Vocabk(c_n)\) & Top-\(k\) subset of the vocabulary for model context \(c_n\). \\
\(\Vocabp(c_n)\) & Top-\(p\), or nucleus, subset of the vocabulary for model context \(c_n\). \\
\bottomrule
\end{longtable}
}

\section{Toy sampler with no AI model}
\label{app:toy-sampler}

Before using an actual language model, it is useful to isolate the sampling mechanism. This example uses no AI model, no prompt, no hidden vectors, and no logits. It simply starts with a fixed probability distribution over four possible tokens:
\[
p=(0.10,0.20,0.30,0.40).
\]
Two pseudo-RNGs initialized with the same seed will produce the same sequence of pseudo-random draws. Therefore, they will produce the same sampled tokens if they consume the same number of draws in the same order.

\begin{lstlisting}[style=pythonstyle]
import torch

probs = torch.tensor([[0.10, 0.20, 0.30, 0.40]])

rng_i = torch.Generator(device="cpu").manual_seed(7)
rng_j = torch.Generator(device="cpu").manual_seed(7)

draws_i = []
draws_j = []

for _ in range(10):
    draws_i.append(torch.multinomial(probs, 1, generator=rng_i).item())
    draws_j.append(torch.multinomial(probs, 1, generator=rng_j).item())

print(draws_i)
print(draws_j)
print("Identical:", draws_i == draws_j)
\end{lstlisting}

The important point is not the particular numbers printed. The important point is that stochastic sampling can be reproducible when the probability distribution, pseudo-RNG state, and sampling procedure are matched.

\section{Twin GPT-2 demonstrations: sampling and greedy decoding}
\label{app:twin-demo}

This appendix demonstrates two related claims. First, stochastic token generation can be reproducible under matched model, input, decoder, and pseudo-RNG conditions. Second, greedy decoding can produce identical twin outputs without using token-level randomness. Both demonstrations use \texttt{openai-community/gpt2}, the small GPT-2 model available through Hugging Face \citep{hf-gpt2}. GPT-2 is more fluent than a minimal testing model while remaining tractable enough for a short Google Colab or CPU demonstration.

The demonstrations use two replicas rather than one model run twice for a reason. The point is to dramatize reproducibility across matched instances, analogous to two servers that have the same model, same input, same decoding rule, and, for sampling, the same pseudo-random stream. A single model run twice can also be reproducible, but only if all relevant state is reset exactly.

The demonstrations run on CPU to reduce hardware-specific nondeterminism and to make the examples portable. Production LLM inference typically runs on GPUs or specialized accelerators, where parallel numerical routines, batching, and execution details can introduce additional reproducibility issues.

\subsection{Reproducible stochastic sampling with twin GPT-2 models}

When this script is run in the specified environment with CPU execution and matching model, tokenizer, and library versions, the two independently loaded GPT-2 instances should produce identical token sequences. If the assertion fails on another machine, likely causes include an environment mismatch, a different model or tokenizer revision, GPU rather than CPU execution, or a backend operation whose deterministic behavior differs across platform or library versions. In that case, restart the runtime, ensure that the script is running on CPU, and verify the model, tokenizer, and library versions.

This first script keeps sampling. It does not turn sampling off. Instead, it makes sampling reproducible by giving each replica its own pseudo-RNG initialized to the same seed. The decoder uses top-\(k\) sampling, so it samples only from the top 50 candidates rather than from the full vocabulary.

\begin{lstlisting}[style=pythonstyle]
# ============================================================
# Twin GPT-2 demonstration with reproducible stochastic sampling
# ============================================================

# In Google Colab, run this once if needed:
# !pip install -q transformers accelerate

import os
import random
import numpy as np
import torch
from transformers import AutoTokenizer, AutoModelForCausalLM

# -----------------------------
# Configuration
# -----------------------------
MODEL_NAME = "openai-community/gpt2"
PROMPT = "Agentic AI systems are stochastic because"
SEED = 12345
STEPS = 40
TEMPERATURE = 0.8   # Must be > 0. T = 0 makes this division undefined (inf/nan); for greedy (T = 0) decoding, use the greedy script below.
TOP_K = 50
DEVICE = "cpu"   # CPU gives the cleanest reproducibility demonstration.

os.environ["TOKENIZERS_PARALLELISM"] = "false"

# -----------------------------
# Reduce avoidable nondeterminism
# -----------------------------
random.seed(SEED)
np.random.seed(SEED)
torch.manual_seed(SEED)
torch.set_num_threads(1)

try:
    torch.use_deterministic_algorithms(True)
except Exception as e:
    print("Could not enable deterministic algorithms:", e)

# -----------------------------
# Load tokenizer and twin models
# -----------------------------
tokenizer = AutoTokenizer.from_pretrained(MODEL_NAME)

model_i = AutoModelForCausalLM.from_pretrained(MODEL_NAME).to(DEVICE).eval()
model_j = AutoModelForCausalLM.from_pretrained(MODEL_NAME).to(DEVICE).eval()

# .eval() matters because it disables training-time randomness such as dropout.

# -----------------------------
# Same initial input sequence
# -----------------------------
input_i = tokenizer(PROMPT, return_tensors="pt").input_ids.to(DEVICE)
input_j = tokenizer(PROMPT, return_tensors="pt").input_ids.to(DEVICE)

# -----------------------------
# Independent pseudo-RNGs with the same seed
# -----------------------------
rng_i = torch.Generator(device=DEVICE).manual_seed(SEED)
rng_j = torch.Generator(device=DEVICE).manual_seed(SEED)

# -----------------------------
# One-token top-k sampling function
# -----------------------------
@torch.inference_mode()
def sample_one_token(model, input_ids, rng):
    """
    Compute next-token logits, restrict to the top-k tokens,
    convert to probabilities, and sample one token using rng.
    """
    output = model(input_ids)

    # Raw next-token scores for the final position in the sequence.
    logits = output.logits[:, -1, :]

    # Temperature reshapes the distribution.
    logits = logits / TEMPERATURE

    # Restrict to the TOP_K highest-scoring tokens.
    top_values, top_indices = torch.topk(logits, k=TOP_K, dim=-1)

    # Convert the restricted logits into probabilities.
    probs = torch.softmax(top_values, dim=-1)

    # Sample a position inside the top-k set using the explicit pseudo-RNG.
    sampled_position = torch.multinomial(
        probs,
        num_samples=1,
        generator=rng,
    )

    # Map the sampled top-k position back to the original vocabulary token ID.
    next_token = top_indices.gather(dim=-1, index=sampled_position)
    return next_token

# -----------------------------
# Generate from both replicas
# -----------------------------
sampled_i = []
sampled_j = []

for n in range(STEPS):
    # Same context at the beginning of each step.
    assert torch.equal(input_i, input_j), f"Contexts differ before step {n}"

    # Pedagogical choice: we recompute the full forward pass each step (stateless,
    # matches the math). This omits the KV cache, so the model reprocesses the whole
    # context every step; production systems cache attention keys/values to avoid
    # that redundant recomputation.
    next_i = sample_one_token(model_i, input_i, rng_i)
    next_j = sample_one_token(model_j, input_j, rng_j)

    # Same model state, same context, same decoder, same RNG state.
    assert torch.equal(next_i, next_j), f"Diverged at step {n}"

    sampled_i.append(next_i.item())
    sampled_j.append(next_j.item())

    # Append the sampled token to each context.
    input_i = torch.cat([input_i, next_i], dim=-1)
    input_j = torch.cat([input_j, next_j], dim=-1)

# -----------------------------
# Results
# -----------------------------
print("Token IDs from model i:")
print(sampled_i)

print("\nToken IDs from model j:")
print(sampled_j)

print("\nSequences identical:", sampled_i == sampled_j)

print("\nOutput from model i:")
print(tokenizer.decode(input_i[0], skip_special_tokens=True))

print("\nOutput from model j:")
print(tokenizer.decode(input_j[0], skip_special_tokens=True))
\end{lstlisting}

The expected diagnostic line is:

\begin{lstlisting}[style=shellstyle]
Sequences identical: True
\end{lstlisting}

The exact generated prose can change if package versions, model revisions, or numerical settings change. The invariant demonstrated by the script is that the two replicas match each other under the same controlled conditions. Sampling is still present: the key line is \texttt{torch.multinomial(..., generator=rng)}. The pseudo-RNG stream is simply synchronized across replicas.

\subsection{Greedy decoding with twin GPT-2 models}

The second script removes token-level sampling entirely. The script makes no \texttt{torch.\allowbreak multinomial} call, and it does not use the pseudo-RNG for token selection. At each step, the decoder chooses the token with the largest logit.

\begin{lstlisting}[style=pythonstyle]
# ============================================================
# Twin GPT-2 demonstration with greedy decoding
# ============================================================

# In Google Colab, run this once if needed:
# !pip install -q transformers accelerate

import os
import random
import numpy as np
import torch
from transformers import AutoTokenizer, AutoModelForCausalLM

# -----------------------------
# Configuration
# -----------------------------
MODEL_NAME = "openai-community/gpt2"
PROMPT = "Agentic AI systems are stochastic because"
SEED = 12345
STEPS = 40
DEVICE = "cpu"   # CPU gives the cleanest reproducibility demonstration.

os.environ["TOKENIZERS_PARALLELISM"] = "false"

# -----------------------------
# Reduce avoidable nondeterminism
# -----------------------------
# These seeds are not used by greedy token selection. They are included
# to keep the broader numerical and software environment controlled.
random.seed(SEED)
np.random.seed(SEED)
torch.manual_seed(SEED)
torch.set_num_threads(1)

try:
    torch.use_deterministic_algorithms(True)
except Exception as e:
    print("Could not enable deterministic algorithms:", e)

# -----------------------------
# Load tokenizer and twin models
# -----------------------------
tokenizer = AutoTokenizer.from_pretrained(MODEL_NAME)

model_i = AutoModelForCausalLM.from_pretrained(MODEL_NAME).to(DEVICE).eval()
model_j = AutoModelForCausalLM.from_pretrained(MODEL_NAME).to(DEVICE).eval()

# .eval() disables training-time randomness such as dropout.

# -----------------------------
# Same initial input sequence
# -----------------------------
input_i = tokenizer(PROMPT, return_tensors="pt").input_ids.to(DEVICE)
input_j = tokenizer(PROMPT, return_tensors="pt").input_ids.to(DEVICE)

# -----------------------------
# Greedy one-token decoder
# -----------------------------
@torch.inference_mode()
def greedy_one_token(model, input_ids):
    """
    Compute next-token logits and choose the highest-scoring token.
    This is deterministic greedy decoding.
    """
    output = model(input_ids)
    logits = output.logits[:, -1, :]
    next_token = torch.argmax(logits, dim=-1, keepdim=True)
    return next_token

# -----------------------------
# Generate from both replicas
# -----------------------------
generated_i = []
generated_j = []

for n in range(STEPS):
    # Same context at the beginning of each step.
    assert torch.equal(input_i, input_j), f"Contexts differ before step {n}"

    # Pedagogical choice: we recompute the full forward pass each step (stateless,
    # matches the math). This omits the KV cache, so the model reprocesses the whole
    # context every step; production systems cache attention keys/values to avoid
    # that redundant recomputation.
    next_i = greedy_one_token(model_i, input_i)
    next_j = greedy_one_token(model_j, input_j)

    # Same model state, same context, same deterministic decoder.
    assert torch.equal(next_i, next_j), f"Diverged at step {n}"

    generated_i.append(next_i.item())
    generated_j.append(next_j.item())

    # Append the selected token to each context.
    input_i = torch.cat([input_i, next_i], dim=-1)
    input_j = torch.cat([input_j, next_j], dim=-1)

# -----------------------------
# Results
# -----------------------------
print("Token IDs from model i:")
print(generated_i)

print("\nToken IDs from model j:")
print(generated_j)

print("\nSequences identical:", generated_i == generated_j)

print("\nOutput from model i:")
print(tokenizer.decode(input_i[0], skip_special_tokens=True))

print("\nOutput from model j:")
print(tokenizer.decode(input_j[0], skip_special_tokens=True))
\end{lstlisting}

The expected diagnostic line is again:

\begin{lstlisting}[style=shellstyle]
Sequences identical: True
\end{lstlisting}

The interpretation is different. In the sampling script, the two replicas match because the stochastic sampling draws are reproducible under synchronized pseudo-RNG states. In the greedy script, the two replicas match because token selection is deterministic: at each step the decoder chooses the highest-scoring token. The greedy output may still be repetitive, bland, or caught in a high-probability loop. That is not a failure of the demonstration. It illustrates that determinism and output quality are separate properties. This is one reason practical systems often use sampling, top-\(k\), top-\(p\), temperature, repetition penalties, or related decoding controls.

\subsection{Why the same visible prompt may not repeat under sampling}

For sampling-based decoding, using the same visible prompt is not sufficient to reproduce a sampled token if the pseudo-RNG state has advanced. To make this effect easy to see, the following snippet uses a high temperature, which flattens the next-token probability distribution and gives lower-ranked tokens more probability mass. This temperature is deliberately exaggerated for demonstration; it is not meant as a recommended production setting.

The point is not that every pair of sampled tokens must differ. Two different pseudo-random draws can still select the same token, especially when the next-token distribution is concentrated. The point is that each sampling call consumes the next pseudo-random draw in the stream. Therefore, to reproduce a sampled draw exactly, one must reset or restore the RNG state, not merely reuse the same visible prompt.

In the example below, the visible prompt remains unchanged across calls, but the pseudo-RNG state advances. With the distribution flattened by a high temperature, the repeated draws often produce different sampled token IDs.

\begin{lstlisting}[style=pythonstyle]
@torch.inference_mode()
def sample_one_token_high_temperature(model, input_ids, rng, temperature):
    """
    Sample one token from the full vocabulary after applying
    a deliberately high temperature for demonstration.
    """
    output = model(input_ids)
    logits = output.logits[:, -1, :] / temperature
    probs = torch.softmax(logits, dim=-1)
    return torch.multinomial(probs, num_samples=1, generator=rng)

input_ids = tokenizer(PROMPT, return_tensors="pt").input_ids.to(DEVICE)

rng = torch.Generator(device=DEVICE).manual_seed(SEED)

TEMPERATURE_DEMO = 5.0
sampled_ids = []

for _ in range(10):
    sampled = sample_one_token_high_temperature(
        model_i,
        input_ids,
        rng,
        temperature=TEMPERATURE_DEMO,
    )
    sampled_ids.append(sampled.item())

sampled_tokens = [tokenizer.decode([token_id]) for token_id in sampled_ids]

print("Sampled token IDs:", sampled_ids)
print("Sampled tokens:", sampled_tokens)
print("Number of distinct sampled token IDs:", len(set(sampled_ids)))
\end{lstlisting}

To verify that the stochastic process is still reproducible, reset the generator and repeat the same ten draws:

\begin{lstlisting}[style=pythonstyle]
rng_replay = torch.Generator(device=DEVICE).manual_seed(SEED)

replayed_ids = []

for _ in range(10):
    sampled = sample_one_token_high_temperature(
        model_i,
        input_ids,
        rng_replay,
        temperature=TEMPERATURE_DEMO,
    )
    replayed_ids.append(sampled.item())

print("Replayed token IDs:", replayed_ids)
print("Replay matches:", sampled_ids == replayed_ids)
\end{lstlisting}

The varied token IDs do not show that the model weights or prompt changed. They show that sampling is a function of both the next-token probability distribution and the pseudo-RNG state. If the generator is reset to the same seed and the same sequence of sampling calls is repeated under the same model and numerical conditions, the same sequence of sampled token IDs should be reproduced.

\subsection{Proof idea for the twin sampling demonstration}

The sampling demonstration rests on an induction argument under controlled numerical conditions.

\begin{enumerate}[leftmargin=*]
\item At step \(0\), both replicas have the same context and the same RNG state.
\item If the model contexts are equal, the deterministic forward passes produce equal logits and equal probabilities.
\item If the RNG states are equal, the samplers receive the same pseudo-random draw.
\item Equal probabilities plus equal pseudo-random draw imply equal selected token.
\item Appending equal tokens preserves equality of contexts, and consuming one draw advances both RNGs in the same way.
\item Therefore, the same reasoning applies at the next step.
\end{enumerate}

The phrase ``deterministic forward pass'' is doing real work. The demonstrations use CPU execution, fixed model weights, evaluation mode, a fixed tokenizer, fixed input shapes, and manual one-token decoding. A production inference server may violate that assumption if batch composition, kernels, precision, routing, or model version differ.

\section{Practical cautions for extending the demonstration}
\label{app:cautions}

The GPT-2 demonstrations are intentionally controlled. Larger models and deployed agentic systems may require more care.

\begin{enumerate}[leftmargin=*]
\item \textbf{Use CPU for the cleanest classroom demonstration.} GPU inference is fast, but exact reproducibility can require deterministic kernels, fixed shapes, fixed batch composition, and batch-invariant serving.

\item \textbf{Use evaluation mode.} Training mode may include dropout or other training-time stochasticity. Inference demonstrations should use \texttt{model.\allowbreak eval()}.

\item \textbf{Avoid hidden random draws.} High-level generation APIs may consume random numbers in ways that are not visible. Manual one-token sampling makes the pseudo-RNG step explicit.

\item \textbf{Keep the RNG streams synchronized.} Same seed at initialization is not enough if one replica consumes an extra random draw. The key condition is same RNG state at each token step.

\item \textbf{Record model and software versions.} Exact output strings can change if the model weights, tokenizer, library version, or numerical kernels change.

\item \textbf{Do not generalize from a toy model to all deployed services.} A commercial agent may include routing, retrieval, tool calls, safety layers, dynamic batching, retries, and changing external data. Those layers can introduce additional variation even if the underlying model call is controlled.
\end{enumerate}

\bibliographystyle{plainnat}
\bibliography{\jobname}

\end{document}